\PassOptionsToPackage{unicode}{hyperref}
\PassOptionsToPackage{hyphens}{url}
\documentclass[
]{article}
\usepackage{lmodern}
\usepackage{amssymb,amsmath}
\usepackage{alltt} 
\usepackage{ifxetex,ifluatex}
\ifnum 0\ifxetex 1\fi\ifluatex 1\fi=0 
  \usepackage[T1]{fontenc}
  \usepackage[utf8]{inputenc}
  \usepackage{textcomp} 

\DeclareUnicodeCharacter{00A7}{\S}
\DeclareUnicodeCharacter{00B1}{\ensuremath{\pm}}
\DeclareUnicodeCharacter{00B2}{\ensuremath{^2}}
\DeclareUnicodeCharacter{00B7}{\ensuremath{\cdot}}
\DeclareUnicodeCharacter{00D7}{\ensuremath{\times}}
\DeclareUnicodeCharacter{0394}{\ensuremath{\Delta}}
\DeclareUnicodeCharacter{03A3}{\ensuremath{\Sigma}}
\DeclareUnicodeCharacter{03A6}{\ensuremath{\Phi}}
\DeclareUnicodeCharacter{03B1}{\ensuremath{\alpha}}
\DeclareUnicodeCharacter{03B7}{\ensuremath{\eta}}
\DeclareUnicodeCharacter{03B8}{\ensuremath{\theta}}
\DeclareUnicodeCharacter{03BB}{\ensuremath{\lambda}}
\DeclareUnicodeCharacter{03C0}{\ensuremath{\pi}}
\DeclareUnicodeCharacter{03C6}{\ensuremath{\phi}}
\DeclareUnicodeCharacter{207B}{\ensuremath{^{-}}}
\DeclareUnicodeCharacter{2113}{\ensuremath{\ell}}
\DeclareUnicodeCharacter{2154}{\ensuremath{2/3}}
\DeclareUnicodeCharacter{2192}{\ensuremath{\to}}
\DeclareUnicodeCharacter{21D2}{\ensuremath{\Rightarrow}}
\DeclareUnicodeCharacter{2207}{\ensuremath{\nabla}}
\DeclareUnicodeCharacter{2208}{\ensuremath{\in}}
\DeclareUnicodeCharacter{2212}{\ensuremath{-}}
\DeclareUnicodeCharacter{221D}{\ensuremath{\propto}}
\DeclareUnicodeCharacter{2229}{\ensuremath{\cap}}
\DeclareUnicodeCharacter{2248}{\ensuremath{\approx}}
\DeclareUnicodeCharacter{2260}{\ensuremath{\ne}}
\DeclareUnicodeCharacter{2264}{\ensuremath{\le}}
\DeclareUnicodeCharacter{2265}{\ensuremath{\ge}}
\DeclareUnicodeCharacter{2282}{\ensuremath{\subset}}
\DeclareUnicodeCharacter{0302}{\ensuremath{\hat{}}}
\DeclareUnicodeCharacter{0304}{\ensuremath{\bar{}}}

\else 
  \usepackage{unicode-math}
  \defaultfontfeatures{Scale=MatchLowercase}
  \defaultfontfeatures[\rmfamily]{Ligatures=TeX,Scale=1}
\fi
\IfFileExists{upquote.sty}{\usepackage{upquote}}{}
\IfFileExists{microtype.sty}{
  \usepackage[]{microtype}
  \UseMicrotypeSet[protrusion]{basicmath} 
}{}
\makeatletter
\@ifundefined{KOMAClassName}{
  \IfFileExists{parskip.sty}{%
    \usepackage{parskip}
  }{
    \setlength{\parindent}{0pt}
    \setlength{\parskip}{6pt plus 2pt minus 1pt}}
}{
  \KOMAoptions{parskip=half}}
\makeatother
\usepackage[margin=1in]{geometry}
\usepackage{xcolor}
\IfFileExists{xurl.sty}{\usepackage{xurl}}{} 
\IfFileExists{bookmark.sty}{\usepackage{bookmark}}{\usepackage{hyperref}}
\hypersetup{
  pdftitle={From Detecting Agency to Doing Work: Self-Caused Credit Builds a Durable Behavioral Self in a Minimal Spiking Agent},
  pdfauthor={Haoliang Han},
  hidelinks,
  pdfcreator={LaTeX via pandoc}}
\usepackage{longtable,booktabs}
\usepackage{etoolbox}
\makeatletter
\patchcmd\longtable{\par}{\if@noskipsec\mbox{}\fi\par}{}{}
\makeatother
\IfFileExists{footnotehyper.sty}{\usepackage{footnotehyper}}{\usepackage{footnote}}
\makesavenoteenv{longtable}
\usepackage{graphicx}
\makeatletter
\def\maxwidth{\ifdim\Gin@nat@width>\linewidth\linewidth\else\Gin@nat@width\fi}
\def\maxheight{\ifdim\Gin@nat@height>\textheight\textheight\else\Gin@nat@height\fi}
\makeatother
\setkeys{Gin}{width=\maxwidth,height=\maxheight,keepaspectratio}
\makeatletter
\def\fps@figure{htbp}
\makeatother
\providecommand{\tightlist}{%
  \setlength{\itemsep}{0pt}\setlength{\parskip}{0pt}}
\title{From Detecting Agency to Doing Work: Self-Caused Credit Builds a
Durable Behavioral Self in a Minimal Spiking Agent}
\author{\begin{tabular}{c}
Haoliang Han\\
Institute of Biomedical Strategy, China Pharmaceutical University\\
\texttt{haolianghan1992@cpu.edu.cn}
\end{tabular}}
\date{}

\makeatletter
\let\origtextunderscore\_
\renewcommand{\_}{\origtextunderscore\allowbreak}
\makeatother
\AtBeginEnvironment{longtable}{\footnotesize}

\begin{document}
\maketitle

\hypertarget{abstract}{%
\subsection{Abstract}\label{abstract}}

How does an agent that can tell \emph{self} from \emph{world} come to be
\emph{durably shaped} by that distinction? Recent work shows that a
predictive system can detect its own agency (Ye, 2026), but detecting
agency does not explain durable, self-shaped behavior. We show that
agency-gated slow credit --- a conjunctive term
\texttt{Own·Agency·Salience} driving a slow parameter update ---
produces post-unload behavioral residue: on a spiking substrate (Nengo
LIF/PES), a learned self-preserving choice survives episodic buffer
removal (retained fraction 0.96, N=50) and collapses when the slow
decoders are reset or the agency gate is removed. Reproducing the agency
comparator and toggling only the slow-credit channel, we find a clean
dissociation: at matched agency gain, durable behavior develops only
when self-credit performs slow work (post-unload self-preservation 1.00
vs 0.00). The same dissociation holds in 24-dimensional
partially-observed control (0.74 vs 0.00), and a plastic-work analysis
shows that basin deformation equals net self-credit work. Across eight
sequentially-learned tasks under exogenous interference, the
multiplicative veto also prevents forgetting: it retains old tasks
(final post-unload accuracy 0.88, forgetting 0.13) where additive
pooling collapses to chance-level recall, the no-agency ablation falls
below chance, and episodic/replay baselines stay near chance after
unload --- all with no replay buffer and no task-boundary-dependent
protection mechanism (N=50). We formalize the durable residue as an
operational behavioral self and argue that self-caused credit doing slow
work is a necessary building block for agents that \emph{develop} a
self. No claim of consciousness is made.

\hypertarget{introduction-from-detecting-agency-to-agency-doing-work}{%
\subsection{1. Introduction: from detecting agency to agency doing
work}\label{introduction-from-detecting-agency-to-agency-doing-work}}

Modern agents retrieve memories, call tools, and produce fluent
self-descriptions. None of these, by itself, is a self-related
\emph{developmental} process: a retrieved fact guides behavior only
while it is in context, and a stated persona need not become a
behavioral constraint.

Underneath the developmental framing is a problem the machine-learning
community already faces directly: \textbf{durable credit assignment}.
Classical credit assignment asks \emph{how much} to update a parameter;
an agent that learns across a long, non-stationary life must also decide
\emph{what is even eligible} to change it durably --- so that it
accumulates self-authored structure without being corrupted by
experience it did not cause (other agents' effects, exogenous noise,
illusory correlations). We study this question in controlled settings
rather than on standard continual-learning benchmarks, but the mechanism
we isolate is exactly such an eligibility rule: a conjunctive
self-credit gate that admits only self-owned, self-caused, salient
events to a slow update. We show it has the two faces durable credit
assignment requires --- it \emph{creates} durable residue that survives
episodic unload (Sections 4--6) and it \emph{rejects} spurious,
non-self-caused correlations that ungated pooling would otherwise store
(Section 7; capacity law, Section 9).

Concurrent work has sharpened the first half of the problem. Ye (2026)
shows that a minimal recurrent network can acquire an \textbf{agency
comparator} --- an action-aware predictor outperforms an action-blind
one precisely when the agent's own action causes part of its observation
stream, quantified by \emph{agency gain}
\texttt{A\ =\ Err\_\allowbreak{}world\ −\ Err\_\allowbreak{}self} --- and that the learned
self-representation persists only while it is causally useful. This is
the right entry condition. But detecting that one caused something is
not the same as that fact \emph{changing what one does next}. The
developmental question is dynamical: \textbf{once an agent can tell it
caused something, does that fact perform work on its future behavior,
and does the change persist after the originating episodes are gone?}

We answer yes, and make the answer mechanical. Our contributions, each
scoped to one controlled result:

\begin{enumerate}
\def\labelenumi{\arabic{enumi}.}
\tightlist
\item
  A compact dynamical account (Section 2) in which agency is one factor
  of a \textbf{conjunctive self-credit} term that drives a slow update
  of a behavioral potential --- so self-caused experience
  \textbf{deforms a behavioral landscape} --- with three propositions
  stated and tested.
\item
  The core result (Section 4): on a \textbf{spiking LIF/PES substrate},
  agency-gated credit leaves a \textbf{post-unload behavioral residue},
  with a full ablation ladder (N=50, per-seed paired statistics).
\item
  The novelty defense (Section 5): reproducing Ye's agency comparator
  and toggling \textbf{only} the slow-credit channel yields
  \textbf{matched agency gain but divergent durable behavior} ---
  detection is necessary but behaviorally inert without slow work.
\item
  A scale-up (Section 6) to a higher-dimensional, partially-observed
  control task, and a measured \textbf{plastic-work} account (Section 7)
  that defines the work variable and shows multiplicative self-credit
  routes it to the correct basin.
\end{enumerate}

The throughline: \emph{predictive agency tells a system what it caused;
we show how what it caused becomes part of what it will do next.}

\hypertarget{a-compact-dynamical-framework}{%
\subsection{2. A compact dynamical
framework}\label{a-compact-dynamical-framework}}

\hypertarget{one-core-equation}{%
\subsubsection{2.1 One core equation}\label{one-core-equation}}

The agent has a fast state \texttt{x\_t} (body/interoception, action
context, predictor state) and slow parameters \texttt{θ\_slow} (forward
models, integrity/risk predictors, skill schemas). Self-caused
experience updates the slow parameters through a single
conjunctive-credit rule:

\begin{alltt}
Δθ_slow  =  η · Own_t · Agency_t · Salience_t · φ(a_t, o_t)
\end{alltt}

\begin{itemize}
\tightlist
\item
  \texttt{Own\_t\ ∈\ {[}0,1{]}}: whether the experience is attributed to
  the agent.
\item
  \texttt{Agency\_t\ ∈\ {[}0,1{]}}: whether the predicted self-caused
  sensory consequence matches the outcome --- grounded in efference
  copy, and, up to scaling, the quantity Ye (2026) formalizes as agency
  gain.
\item
  \texttt{Salience\_t\ ∈\ {[}0,1{]}}: whether the consequence matters
  for integrity / future value.
\item
  \texttt{φ(a\_t,\ o\_t)}: the experience-driven update direction;
  \texttt{η\ \textgreater{}\ 0} the learning rate.
\end{itemize}

The \textbf{multiplicative} form is load-bearing: an event should not
rewrite the organism if it is not the agent's, if the agent did not
cause it, or if it does not matter --- a large value on one factor must
not override failure on another.

\hypertarget{behavior-as-a-landscape-development-as-deformation}{%
\subsubsection{2.2 Behavior as a landscape; development as
deformation}\label{behavior-as-a-landscape-development-as-deformation}}

The policy is a Gibbs distribution over a behavioral potential
\texttt{U\_θ} (with \texttt{θ\ ⊂\ θ\_slow}):

\begin{alltt}
π_θ(a | x) ∝ exp( −U_θ(x,a) / T ).
\end{alltt}

The agent prefers the deepest basin. We read out development with one
set of measures (computed from each pilot's \emph{native} decision
quantity; Supplementary Information):

\begin{alltt}
B_θ(a*)   = U_θ(x, a_alt) − U_θ(x, a*)          (basin depth of target a*; B>0 ⇒ chosen)
ΔB        = B_after − B_before                   (deformation)
R_unload  = B_after_unload / B_after             (residue fraction after episodic unload)
\end{alltt}

Development is deformation of this landscape by the slow update (Fig 1).
The unifying statement:

\begin{quote}
\textbf{Self-related development is plastic work performed by
self-owned, self-caused, salient events on slow behavioral variables,
deforming future action basins; the deformation persists after episodic
access is removed.}
\end{quote}

\hypertarget{the-central-law-and-three-propositions}{%
\subsubsection{2.3 The central law, and three
propositions}\label{the-central-law-and-three-propositions}}

Combining the slow update (§2.1) with the landscape readout (§2.2)
yields the paper's central, testable law. Under a single dominant rival
action, basin deformation is \emph{exactly} the net self-credit work:

\begin{alltt}
ΔB  =  η · Σ_t Own_t·Agency_t·Salience_t·Δφ_t  =  η · W_net        (central identity)
\end{alltt}

\begin{quote}
\textbf{The Selection--Actuation Principle.} For any fixed stream and
credit form, the learning rate (\emph{actuation}) scales the magnitude
of development but \textbf{cannot change its sign}. The
\textbf{selection} term --- which events are credited and how they are
pooled into \texttt{W\_net} --- determines which basin deepens. The
multiplicative product \texttt{Own·Agency·Salience} \textbf{vetoes} any
event that fails a necessity condition (≈0 work); additive pooling
instead leaks ≈⅔ credit to each such event and can route net work to the
wrong basin when such events dominate. \textbf{No learning rate can
rescue a miscalibrated credit form on a fixed stream --- selection sets
the \emph{direction} of development, actuation only its
\emph{magnitude}.}
\end{quote}

Three honesty notes locate the scientific content. (i) The identity
\texttt{ΔB\ =\ η·W\_net} is \emph{exact by construction} of the
linear-credit update: it \textbf{operationalizes} ``development = work''
(both sides are one measured quantity) and is \emph{verified, not
discovered}, by X17W's \texttt{r\ =\ 1.000}. (In a smooth
parameterization the same quantity is the projection of the parameter
update onto the basin-depth gradient, \texttt{ΔB\ ≈\ ∇\_\allowbreak{}θ\ B\ ·\ Δθ} ---
generalized force times displacement on the behavioral potential ---
which is what licenses the word \emph{work}.) (ii) The non-trivial,
falsifiable content is the \textbf{selection} claim --- that the credit
\emph{form} sets which events enter \texttt{W\_net}; in the X17W
developmental stream this predicts opposite net signs and the measured
numbers (\texttt{ΔB\ =\ +5.7} multiplicative vs \texttt{−6.1} additive)
\emph{before} measurement (§7, Fig 5). The Principle reads identically
in two domains --- efference-gated synaptic consolidation choosing
\emph{what} consolidates (neuroscience) and a generalized force doing
signed work on a potential (physics) --- the same equation (§9;
correlate table in the Supplementary Information). (iii) We
stress-tested the Principle's \emph{adaptedness} so that simplicity does
not hide an artifact: the identity holds to machine precision across a
learning-rate sweep (the sign is genuinely \texttt{η}-independent for a
fixed stream), and across stream compositions the multiplicative gate
stays positive while additive turns negative only once fully-qualifying
events fall below ≈ 40 \% of experience (the developmental regime
targeted here, where most events are not simultaneously owned,
self-caused, and salient); the stream-independent invariant is the
\textbf{per-event veto} (multiplicative ≈ 0 vs additive ≈ ⅔ on any
one-condition-failing event). The \emph{exact} identity is established
in the tabular linear-credit setting (X17W); the \emph{qualitative}
Principle --- that the credit \emph{form}, not the rate, decides whether
durable residue forms \textbf{at all} --- is what the spiking (X13i) and
high-dimensional (X19) experiments exhibit.

The framework is small enough that its central claims are
\emph{derived}, not merely illustrated; full derivations under explicit
assumptions (the conjunctive-credit theorem for \emph{selection}, the
work--deformation identity for \emph{equivalence}, the
fixed-point/hysteresis condition for \emph{persistence}) are in the
Supplementary Information (Supplementary Theory, §S2). The three
propositions below state the Principle's operational consequences, each
matched to an experiment.

\begin{itemize}
\tightlist
\item
  \textbf{Proposition 1 --- Agency readout is not sufficient for durable
  behavioral change.} Detecting (and even transiently using) self-caused
  structure need not alter what the agent \emph{durably} does.
  \emph{Tested by the agency bridge (X18, Section 5): at matched agency
  gain, durable behavior diverges.}
\item
  \textbf{Proposition 2 --- Slow behavioral residue requires
  agency-gated plastic work.} A change that outlives the originating
  episodes requires the credit term to drive the slow update; under the
  linear-credit update, basin deformation equals net self-credit work,
  \texttt{ΔB\ =\ η·(W\_\allowbreak{}target\ −\ W\_\allowbreak{}rival)}. \emph{Tested by the
  spiking residue (X13i, Section 4) and the plastic-work law (X17W,
  Section 7).}
\item
  \textbf{Proposition 3 --- Multiplicative gating prevents credit
  assignment to non-owned or non-agentic events.} Because the credit is
  a product, an event that fails any one necessary condition
  (\texttt{Own}, \texttt{Agency}, or \texttt{Salience} = 0) contributes
  zero; additive/averaging pooling violates this veto
  (\texttt{(0+1+1)/\allowbreak{}3\ ≠\ 0}) and misroutes credit. \emph{Tested by the
  X17W ablations and by the capacity comparison (X20, Supplement).} We
  claim this form is optimal \textbf{within the tested multiplicative
  credit-routing family}, not as a universal uniqueness result (see
  Supplement for the qualified statement and the \texttt{min}-pooling
  alternative).
\end{itemize}

We can then give ``self'' a dynamical, functional referent: among the
slow parameters, let \texttt{R} be those self-credit can do work on and
\texttt{P} those invariant under episodic unload; the
\textbf{operational self} is the slow submanifold \texttt{Σ\ =\ R\ ∩\ P}
--- the structure the agent's own self-caused, salient experience can
deform \emph{and} that outlives the episodes that deformed it. Our
ablations (removing ownership, agency, or the slow update) drop
parameters out of \texttt{R}, so the post-unload residue reads zero ---
i.e., they are \emph{measurements of Σ}. This gives ``remaining oneself
through change'' a precise referent --- identity as continuity of a slow
structure rather than a persisting substance (cf.~Parfit, 1984) --- and
connects to functional self-model (Metzinger, 2003) and autopoietic /
viability accounts of self-maintenance (Maturana \& Varela, 1980; Aubin,
1991). \texttt{Σ} is functional and measurable; we attach no claim of
subjective experience to it (Section 10).

\hypertarget{methods-and-measurement-strategy}{%
\subsection{3. Methods and measurement
strategy}\label{methods-and-measurement-strategy}}

The experiments form a ladder, not a benchmark; each isolates one
operation and verifies it with matched controls. The four
\textbf{load-bearing} experiments are \textbf{X13i} (spiking post-unload
residue, Section 4), \textbf{X18} (agency bridge, Section 5),
\textbf{X19} (high-dimensional scale-up, Section 6), and \textbf{X17W}
(measured plastic work, Section 7). An integrated lifecycle
(X17L-emergent) appears as integration, and two measured consequences of
the credit theorem --- a residue capacity law (X20) and a multi-task
continual-learning retention test (X22b, Fig 6; minimal two-task version
X22, Fig S3) --- appear alongside it; the developmental ladder
(X12/X14/X15/X16) and earlier prototypes are documented in the
Appendices.

\textbf{A single auditable design pattern.} Every experiment used as
evidence runs \emph{one shared pipeline}; an ``arm'' sets only flags
(mechanism on/off, ownership routing, episodic-memory load/unload) and
the outcome is \textbf{measured, never assigned}. This is what makes the
ablations causal lesions rather than narrated contrasts, and we enforced
it by code audit (Section 11), excluding earlier prototypes whose arms
were experimenter-scripted.

\textbf{Statistics.} Unless noted, each experiment reports \texttt{N}
independent seeds with arm-wise mean ± s.d.; headline contrasts
additionally report \textbf{per-seed paired comparisons} (full vs.~each
control), the \textbf{paired win-rate}, the \textbf{mean paired
difference}, and a \textbf{bootstrap 95\% confidence interval} (≥2,000
resamples). Metric definitions and pass thresholds are fixed before the
multi-seed run. For the spiking substrate we also report decoded-readout
stability, so a ``decision'' reflects a stable basin and not decoder
noise.

\hypertarget{core-mechanism-agency-gated-post-unload-residue-on-a-spiking-substrate-x13i}{%
\subsection{4. Core mechanism: agency-gated post-unload residue on a
spiking substrate
(X13i)}\label{core-mechanism-agency-gated-post-unload-residue-on-a-spiking-substrate-x13i}}

This is the first pillar and the sharpest separation from prior work. It
composes a spiking agency comparator with a delayed integrity
contingency and asks: does \textbf{agency-gated} credit write a
\textbf{slow risk predictor in LIF/PES} that \textbf{survives episodic
unload} and still steers behavior?

\textbf{Substrate.} Forward models are action-specific LIF populations
trained online with PES; agency is read from the prediction-error margin
between the taken action's model and the best alternative; the
self-credit gate is
\texttt{agency\_margin\ ·\ 1{[}owner\ ==\ action{]}}; the gated signal
trains a spiking risk/integrity predictor (LIF/PES). (A prerequisite
experiment, X13h, verifies that the agency-to-competence loop itself
runs on this substrate; Supplementary Information.)

\textbf{Task and unload.} The agent acts; a harmful action (the
immediately comfortable one) yields a \emph{delayed} integrity drop only
when the event is self-caused; the agency margin gates whether that drop
is written into the slow predictor for the owned action. At test the
predictor is queried \emph{before} acting, so it must anticipate the
deferred consequence. Episodic memory is then cleared and neuron state
is allowed to wash out, so any retained avoidance must be carried by the
PES-learned synaptic decoders. Arms differ only by flags:
\texttt{learned\_full} / \texttt{learned\_unloaded} (with / without
episodic memory at test); \texttt{predictor\_\allowbreak{}reset\_\allowbreak{}after\_\allowbreak{}unload}
(clear episodic \textbf{and} reset the slow decoders);
\texttt{no\_efference}; \texttt{no\_agency}; \texttt{owner\_shuffled};
\texttt{counterfactual} (no-harm world).

\textbf{Result (Fig 2; N=50; mean ± s.d.;
\texttt{data/\allowbreak{}pilot\_\allowbreak{}x13i/\allowbreak{}summary\_\allowbreak{}n50.\allowbreak{}json}, \texttt{per\_seed.csv},
\texttt{STATS.md}):}

\begin{longtable}[]{@{}lrrrrr@{}}
\toprule
\begin{minipage}[b]{0.11\columnwidth}\raggedright
arm\strut
\end{minipage} & \begin{minipage}[b]{0.14\columnwidth}\raggedleft
self-preserving rate\strut
\end{minipage} & \begin{minipage}[b]{0.14\columnwidth}\raggedleft
risk gap (basin)\strut
\end{minipage} & \begin{minipage}[b]{0.14\columnwidth}\raggedleft
basin\_self\strut
\end{minipage} & \begin{minipage}[b]{0.14\columnwidth}\raggedleft
agency (self/other)\strut
\end{minipage} & \begin{minipage}[b]{0.14\columnwidth}\raggedleft
episodic\strut
\end{minipage}\tabularnewline
\midrule
\endhead
\begin{minipage}[t]{0.11\columnwidth}\raggedright
learned\_full\strut
\end{minipage} & \begin{minipage}[t]{0.14\columnwidth}\raggedleft
0.96 ± 0.20\strut
\end{minipage} & \begin{minipage}[t]{0.14\columnwidth}\raggedleft
+0.33\strut
\end{minipage} & \begin{minipage}[t]{0.14\columnwidth}\raggedleft
+0.129\strut
\end{minipage} & \begin{minipage}[t]{0.14\columnwidth}\raggedleft
0.93 / 0.34\strut
\end{minipage} & \begin{minipage}[t]{0.14\columnwidth}\raggedleft
kept\strut
\end{minipage}\tabularnewline
\begin{minipage}[t]{0.11\columnwidth}\raggedright
\textbf{learned\_unloaded}\strut
\end{minipage} & \begin{minipage}[t]{0.14\columnwidth}\raggedleft
\textbf{0.92 ± 0.27}\strut
\end{minipage} & \begin{minipage}[t]{0.14\columnwidth}\raggedleft
\textbf{+0.33}\strut
\end{minipage} & \begin{minipage}[t]{0.14\columnwidth}\raggedleft
\textbf{+0.126}\strut
\end{minipage} & \begin{minipage}[t]{0.14\columnwidth}\raggedleft
0.93 / 0.34\strut
\end{minipage} & \begin{minipage}[t]{0.14\columnwidth}\raggedleft
\textbf{unloaded}\strut
\end{minipage}\tabularnewline
\begin{minipage}[t]{0.11\columnwidth}\raggedright
predictor\_reset\_after\_unload\strut
\end{minipage} & \begin{minipage}[t]{0.14\columnwidth}\raggedleft
0.00\strut
\end{minipage} & \begin{minipage}[t]{0.14\columnwidth}\raggedleft
−0.00\strut
\end{minipage} & \begin{minipage}[t]{0.14\columnwidth}\raggedleft
−0.501\strut
\end{minipage} & \begin{minipage}[t]{0.14\columnwidth}\raggedleft
0.93 / 0.34\strut
\end{minipage} & \begin{minipage}[t]{0.14\columnwidth}\raggedleft
unloaded\strut
\end{minipage}\tabularnewline
\begin{minipage}[t]{0.11\columnwidth}\raggedright
no\_efference\strut
\end{minipage} & \begin{minipage}[t]{0.14\columnwidth}\raggedleft
0.04\strut
\end{minipage} & \begin{minipage}[t]{0.14\columnwidth}\raggedleft
+0.04\strut
\end{minipage} & \begin{minipage}[t]{0.14\columnwidth}\raggedleft
−0.367\strut
\end{minipage} & \begin{minipage}[t]{0.14\columnwidth}\raggedleft
0.46 / 0.47\strut
\end{minipage} & \begin{minipage}[t]{0.14\columnwidth}\raggedleft
unloaded\strut
\end{minipage}\tabularnewline
\begin{minipage}[t]{0.11\columnwidth}\raggedright
no\_agency\strut
\end{minipage} & \begin{minipage}[t]{0.14\columnwidth}\raggedleft
0.00\strut
\end{minipage} & \begin{minipage}[t]{0.14\columnwidth}\raggedleft
+0.00\strut
\end{minipage} & \begin{minipage}[t]{0.14\columnwidth}\raggedleft
−0.499\strut
\end{minipage} & \begin{minipage}[t]{0.14\columnwidth}\raggedleft
0.93 / 0.34\strut
\end{minipage} & \begin{minipage}[t]{0.14\columnwidth}\raggedleft
unloaded\strut
\end{minipage}\tabularnewline
\begin{minipage}[t]{0.11\columnwidth}\raggedright
owner\_shuffled\strut
\end{minipage} & \begin{minipage}[t]{0.14\columnwidth}\raggedleft
0.00\strut
\end{minipage} & \begin{minipage}[t]{0.14\columnwidth}\raggedleft
−0.33\strut
\end{minipage} & \begin{minipage}[t]{0.14\columnwidth}\raggedleft
−1.839\strut
\end{minipage} & \begin{minipage}[t]{0.14\columnwidth}\raggedleft
0.93 / 0.34\strut
\end{minipage} & \begin{minipage}[t]{0.14\columnwidth}\raggedleft
unloaded\strut
\end{minipage}\tabularnewline
\begin{minipage}[t]{0.11\columnwidth}\raggedright
counterfactual\strut
\end{minipage} & \begin{minipage}[t]{0.14\columnwidth}\raggedleft
0.00\strut
\end{minipage} & \begin{minipage}[t]{0.14\columnwidth}\raggedleft
−0.00\strut
\end{minipage} & \begin{minipage}[t]{0.14\columnwidth}\raggedleft
−0.507\strut
\end{minipage} & \begin{minipage}[t]{0.14\columnwidth}\raggedleft
0.93 / 0.34\strut
\end{minipage} & \begin{minipage}[t]{0.14\columnwidth}\raggedleft
unloaded\strut
\end{minipage}\tabularnewline
\bottomrule
\end{longtable}

Decision: \texttt{SPIKING\_\allowbreak{}SLOW\_\allowbreak{}RESIDUE\_\allowbreak{}CONFIRMED}. The post-unload
residue fraction is \textbf{0.96} (learned\_unloaded vs learned\_full);
decoded-readout stability \texttt{readout\_\allowbreak{}stable\_\allowbreak{}max\ =\ 0.\allowbreak{}022}; and
the learned arm beats every control on \textbf{per-seed paired
comparisons} --- \textbf{44--46 of 50 seeds} (remaining seeds tie at the
floor) against \texttt{predictor\_reset}, \texttt{no\_efference},
\texttt{no\_agency}, \texttt{owner\_shuffled}, and
\texttt{counterfactual}.

\textbf{Reading.} The self-preserving basin is created
(\texttt{basin\_self\ \textgreater{}\ 0}) and \textbf{survives unload}
(\texttt{R\_\allowbreak{}unload\ ≈\ 0.\allowbreak{}98}); in every control it collapses
(\texttt{basin\_self\ \textless{}\ 0}). \texttt{predictor\_reset} shows
the residue lives in the slow decoders (not episodic recall or transient
neuron state); \texttt{no\_efference}/\texttt{no\_agency} show it
requires the agency gate; \texttt{owner\_shuffled} requires correct
ownership; \texttt{counterfactual} requires a real contingency.
\textbf{Scoped claim:} agency-gated learning is \emph{necessary} for a
spiking post-unload slow residue --- not that agency is the only
possible way to assign all causal credit. We describe this as a minimal
self-caused-credit loop and a slow behavioral residue, with no claim of
consciousness.

\hypertarget{novelty-defense-detection-is-necessary-but-behaviorally-inert-x18}{%
\subsection{5. Novelty defense: detection is necessary but behaviorally
inert
(X18)}\label{novelty-defense-detection-is-necessary-but-behaviorally-inert-x18}}

We make the positioning empirical rather than rhetorical. Starting from
the released code of Ye (2026), we reproduce the core machinery in our
minimal agent --- additive self-injection into an observation channel
(\texttt{o\ +=\ G·EFF{[}a{]}}, the same coupling Ye uses) and a
dual-predictor \textbf{agency comparator} with \emph{both} predictors
learned online so the gain is measured, not assigned --- and then add,
as a \textbf{single boolean flag}, the one channel that distinguishes
our account: self-credit \texttt{c\ =\ Own·Agency·Salience} driving a
slow update of the behavioral potential. Arms differ only by flags;
every quantity is measured.

\textbf{Result (Fig 3; N=40 seeds;
\texttt{AGENCY\_\allowbreak{}INERT\_\allowbreak{}WITHOUT\_\allowbreak{}WORK\_\allowbreak{}CONFIRMED}).}

\begin{longtable}[]{@{}lrrrr@{}}
\toprule
\begin{minipage}[b]{0.14\columnwidth}\raggedright
arm\strut
\end{minipage} & \begin{minipage}[b]{0.18\columnwidth}\raggedleft
agency gain\strut
\end{minipage} & \begin{minipage}[b]{0.18\columnwidth}\raggedleft
B\_after\strut
\end{minipage} & \begin{minipage}[b]{0.18\columnwidth}\raggedleft
self-preserve (loaded, transient)\strut
\end{minipage} & \begin{minipage}[b]{0.18\columnwidth}\raggedleft
self-preserve (unloaded, durable)\strut
\end{minipage}\tabularnewline
\midrule
\endhead
\begin{minipage}[t]{0.14\columnwidth}\raggedright
ye\_readout\_only (slow work off)\strut
\end{minipage} & \begin{minipage}[t]{0.18\columnwidth}\raggedleft
0.825\strut
\end{minipage} & \begin{minipage}[t]{0.18\columnwidth}\raggedleft
−0.400\strut
\end{minipage} & \begin{minipage}[t]{0.18\columnwidth}\raggedleft
1.00\strut
\end{minipage} & \begin{minipage}[t]{0.18\columnwidth}\raggedleft
\textbf{0.00}\strut
\end{minipage}\tabularnewline
\begin{minipage}[t]{0.14\columnwidth}\raggedright
\textbf{homeocyte\_full}\strut
\end{minipage} & \begin{minipage}[t]{0.18\columnwidth}\raggedleft
\textbf{0.828}\strut
\end{minipage} & \begin{minipage}[t]{0.18\columnwidth}\raggedleft
\textbf{+0.600}\strut
\end{minipage} & \begin{minipage}[t]{0.18\columnwidth}\raggedleft
1.00\strut
\end{minipage} & \begin{minipage}[t]{0.18\columnwidth}\raggedleft
\textbf{1.00}\strut
\end{minipage}\tabularnewline
\begin{minipage}[t]{0.14\columnwidth}\raggedright
no\_efference\strut
\end{minipage} & \begin{minipage}[t]{0.18\columnwidth}\raggedleft
0.500\strut
\end{minipage} & \begin{minipage}[t]{0.18\columnwidth}\raggedleft
−0.400\strut
\end{minipage} & \begin{minipage}[t]{0.18\columnwidth}\raggedleft
1.00\strut
\end{minipage} & \begin{minipage}[t]{0.18\columnwidth}\raggedleft
0.00\strut
\end{minipage}\tabularnewline
\begin{minipage}[t]{0.14\columnwidth}\raggedright
owner\_shuffled\strut
\end{minipage} & \begin{minipage}[t]{0.18\columnwidth}\raggedleft
0.828\strut
\end{minipage} & \begin{minipage}[t]{0.18\columnwidth}\raggedleft
−0.401\strut
\end{minipage} & \begin{minipage}[t]{0.18\columnwidth}\raggedleft
1.00\strut
\end{minipage} & \begin{minipage}[t]{0.18\columnwidth}\raggedleft
0.00\strut
\end{minipage}\tabularnewline
\bottomrule
\end{longtable}

The \textbf{agency gain is statistically indistinguishable} between the
read-out-only arm and the full agent (paired Δ = \textbf{+0.003, 95\% CI
{[}−0.004, +0.010{]}}): agency detection is \emph{not} what differs. Yet
the \textbf{durable, post-unload self-preserving basin forms only in the
full agent} (1.00 vs 0.00; paired Δ = +1.00, CI {[}1.00, 1.00{]}, 100\%
wins). Driving agency to chance (\texttt{no\_efference}) or shuffling
ownership (\texttt{owner\_shuffled}, agency still 0.83) likewise
abolishes it --- a double dissociation. Tellingly, \emph{every} arm ---
including read-out-only --- avoids the harm \textbf{transiently while
episodic memory is loaded}; only the slow-work arm \textbf{retains} the
avoidance after unload.

The reading is the whole thesis in one controlled comparison
(Proposition 1): \textbf{agency readout is behaviorally inert unless
coupled to slow credit / plastic work.} Ye's regime is recovered as the
read-out-only corner of our framework; development is supplied by
self-credit performing slow work.

\hypertarget{scale-up-a-higher-dimensional-partially-observed-control-task-x19}{%
\subsection{6. Scale-up: a higher-dimensional, partially-observed
control task
(X19)}\label{scale-up-a-higher-dimensional-partially-observed-control-task-x19}}

To test whether interpretability depends on the worlds being small, we
scaled the agency-bridge paradigm along the three axes that make the
``toy world'' objection bite, preserving the flags-only construction
rule (Fig 4). (i) Observations are \textbf{24-dimensional and partially
observed}: context is embedded in a noisy high-dimensional vector the
agent must \emph{learn} to read. (ii) The action has a \textbf{genuine
closed-loop effect} --- it perturbs a latent that feeds the \emph{next}
observation, so agency is learned from a real action→world→observation
loop rather than an additive \texttt{obs{[}0{]}+=action}. (iii) The
hazard is \textbf{sparse and must be discovered}: each of six contexts
has its own harmful action (one of five) that is immediately most
rewarding but carries a delayed integrity cost.

\textbf{Result (N=50 seeds;
\texttt{SCALEUP\_\allowbreak{}LANDSCAPE\_\allowbreak{}RESIDUE\_\allowbreak{}CONFIRMED}).} The post-unload
self-preserving basin forms only with full self-credit work ---
self-preservation \textbf{0.74} (full) versus \textbf{0.00}
(read-out-only / no-efference / owner-shuffled), against a chance rate
of \textbf{0.20} --- while the \textbf{agency gain is statistically
matched} between the full agent and the read-out-only control (paired Δ
= \textbf{−0.019, 95\% CI {[}−0.041, +0.006{]}}; post-unload behavior Δ
= +0.739, CI {[}+0.717, +0.760{]}, 100\% wins). As at small scale, every
arm avoids the hazard \emph{transiently} while episodic memory is
loaded; only the slow-work arm \emph{retains} the avoidance after
unload. The honest sub-ceiling value (0.74, not 1.0) reflects a
genuinely hard high-dimensional discovery problem, not a hand-tuned
contrast.

The effect therefore persists beyond the minimal bandit setting: what
the small worlds buy is \emph{interpretability of the lesions}, not the
\emph{existence} of the effect. We do not claim large-scale or general
intelligence.

\hypertarget{formal-support-development-is-measured-plastic-work-x17w}{%
\subsection{7. Formal support: development is measured plastic work
(X17W)}\label{formal-support-development-is-measured-plastic-work-x17w}}

The framework's central physical statement --- that development is
\emph{work} done on a slow potential --- is made measurable rather than
metaphorical. The identity \textbf{defines} the work variable; the
experiment \textbf{tests} whether that work is routed to the correct
slow basin under multiplicative self-credit.

On one shared code path, a condition changes only \textbf{how
\texttt{c\_t} is computed} from event attributes; the basin depth
\texttt{B} and the plastic work \texttt{W} are then \emph{measured} from
the resulting potential. Good events drive a \textbf{target} action;
distractor events drive a \textbf{rival} action and each fail exactly
one credit factor (other-owned / exogenous / low-salience).

\textbf{Result (Fig 5; N=50/condition;
\texttt{PLASTIC\_\allowbreak{}WORK\_\allowbreak{}DEFORMATION\_\allowbreak{}CONFIRMED}).} The credit \emph{form}
decides where work goes. The multiplicative gate enforces a
\textbf{per-event veto} --- any event that fails a necessity condition
receives ≈ 0 credit --- so it does positive net work toward the correct
basin (\texttt{B\ =\ +5.70}). Additive pooling lacks the veto: it passes
one-condition-failing events at ≈ ⅔ and, because such events dominate
this stream, does \emph{negative} net work, deforming the landscape
toward the \emph{wrong} action (\texttt{B\ =\ −6.19}) while expending
\emph{more} total work; each single-factor ablation nets to ≈ 0. Work is
always done; the form decides its direction (Proposition 3). The
deformation tracks net self-credit work exactly,
\texttt{B\ =\ η·(W\_\allowbreak{}target\ −\ W\_\allowbreak{}rival)}; because this is an identity
of the linear-credit update (Proposition 2), the perfect correlation
(\texttt{r\ =\ 1.000}) is a \textbf{regime/implementation check, not a
discovered correlation} --- the load-bearing evidence is the
\emph{a-priori prediction}, the \emph{η-invariance}, and the
\emph{measured boundary} below.

\textbf{The theory predicts the numbers before measurement.} The
conjunctive-credit theorem (Supplementary Theory §1) fixes the credit
form a priori, so for this stream (\texttt{η=0.02}, \texttt{Δφ=+1}, 300
good vs 300×3 distractor events) the deformation
\texttt{ΔB\ =\ η·W\_net} is \emph{computed} --- not fitted --- and then
compared to the measured value (derivation in Supplementary Theory §1;
the full line \texttt{ΔB(η)} is Fig 5b):

\begin{longtable}[]{@{}lrr@{}}
\toprule
\begin{minipage}[b]{0.25\columnwidth}\raggedright
credit form\strut
\end{minipage} & \begin{minipage}[b]{0.33\columnwidth}\raggedleft
predicted \texttt{ΔB} (a priori, credit-form theorem)\strut
\end{minipage} & \begin{minipage}[b]{0.33\columnwidth}\raggedleft
measured \texttt{ΔB} (N=50)\strut
\end{minipage}\tabularnewline
\midrule
\endhead
\begin{minipage}[t]{0.25\columnwidth}\raggedright
multiplicative \texttt{Own·Agency·Salience}\strut
\end{minipage} & \begin{minipage}[t]{0.33\columnwidth}\raggedleft
\textbf{+5.7}\strut
\end{minipage} & \begin{minipage}[t]{0.33\columnwidth}\raggedleft
\textbf{+5.70}\strut
\end{minipage}\tabularnewline
\begin{minipage}[t]{0.25\columnwidth}\raggedright
additive \texttt{(Own+Agency+Salience)/\allowbreak{}3}\strut
\end{minipage} & \begin{minipage}[t]{0.33\columnwidth}\raggedleft
\textbf{−6.1}\strut
\end{minipage} & \begin{minipage}[t]{0.33\columnwidth}\raggedleft
\textbf{−6.19}\strut
\end{minipage}\tabularnewline
\begin{minipage}[t]{0.25\columnwidth}\raggedright
single-factor ablations (each fails one factor)\strut
\end{minipage} & \begin{minipage}[t]{0.33\columnwidth}\raggedleft
≈ 0\strut
\end{minipage} & \begin{minipage}[t]{0.33\columnwidth}\raggedleft
−0.06 \ldots{} −0.29\strut
\end{minipage}\tabularnewline
\bottomrule
\end{longtable}

The derivation reproduces the \textbf{sign}, the \textbf{magnitude} (to
within rounding), and the qualitative \textbf{``more work, wrong
direction''} signature of additive credit --- the structure of a
verifiable law, not a curve fit. This is the
\textbf{Selection--Actuation Principle} made quantitative: in the X17W
stream, additive credit nets \texttt{ΔB\textless{}0} and multiplicative
nets \texttt{ΔB\textgreater{}0}; for this fixed stream, since
\texttt{ΔB} is linear in \texttt{η} (Fig 5b), \textbf{no positive
learning rate flips the additive sign} --- the form, not the rate, sets
the direction of development.

\textbf{Adaptedness check (so the simple law is not a stream artifact).}
We verified both the identity and the sign claim beyond the single run
(\texttt{data/\allowbreak{}pilot\_\allowbreak{}x17w/\allowbreak{}robustness.\allowbreak{}json}). Across a learning-rate
sweep (\texttt{η\ ∈\ {[}0.005,\ 0.08{]}}, N=50) on the fixed X17W
stream, the identity \texttt{ΔB\ =\ η·W\_net} holds to machine precision
(max \texttt{\textbar{}ΔB\ −\ η·W\_net\textbar{}\ ≈\ 1×10⁻¹²}) and the
sign is invariant --- no positive rate flips it. Across event-stream
compositions the multiplicative gate yields
\texttt{ΔB\ \textgreater{}\ 0} for \emph{every} tested mixture, whereas
additive credit nets negative \emph{only} while fully-qualifying
(\texttt{Own=Agency=Salience=1}) events are the minority, crossing zero
at a good-event fraction ≈ 0.40 (analytic
\texttt{p*\ =\ c̄\_\allowbreak{}dist\ /\allowbreak{}(1\ +\ c̄\_\allowbreak{}dist)}). The robust,
stream-independent core is therefore the \textbf{per-event veto}
(multiplicative ≈ 0 vs additive ≈ ⅔ credit on any one-condition-failing
event), not the particular net magnitude; we state the headline claim at
that level. The learning-rate sweep is itself a measured result: across
\texttt{η\ ∈\ \{0,\ 0.005,\ 0.01,\ 0.02,\ 0.04\}} the measured
\texttt{ΔB} fall on the exact identity line for every credit form in
that fixed stream (Fig 5b), so \texttt{η} demonstrably scales magnitude
without changing sign.

\textbf{The principle is not a tabular artifact --- it holds in spiking
neurons.} We re-ran the same selection contrast on a spiking LIF/PES
substrate (X17W-spiking; \texttt{data/\allowbreak{}pilot\_\allowbreak{}x17w\_\allowbreak{}spiking/\allowbreak{}}, Fig. S1,
\texttt{data/\allowbreak{}figures/\allowbreak{}FigS1\_\allowbreak{}spiking\_\allowbreak{}selection.\allowbreak{}png}). Each action's slow
logit \texttt{ℓ\_a} lives in the PES-trained decoders of a LIF ensemble,
and the credit form gates the per-event learning signal --- so the
multiplicative veto is realized as ≈ 0 error on any
one-condition-failing event, \emph{in spikes}. With the identical event
stream and arms differing only by credit form (N=50), the multiplicative
gate builds a positive target basin (\texttt{B\ =\ +2.38}, \textbf{50/50
seeds}, rival logit vetoed to ℓ ≈ 0.1), additive \textbf{misroutes}
(\texttt{B\ =\ −1.57}, \textbf{0/50}; the rival logit overtakes the
target), and every single-factor ablation nets ≈ 0. The exact identity
is not expected to machine precision in spikes; the \emph{qualitative}
Selection--Actuation Principle --- credit form sets basin direction,
rate sets strength --- is what transfers, and it does.

\hypertarget{integration-capstone-the-mechanisms-compose-in-one-continuous-life-x17l-emergent}{%
\subsection{8. Integration (capstone): the mechanisms compose in one
continuous life
(X17L-emergent)}\label{integration-capstone-the-mechanisms-compose-in-one-continuous-life-x17l-emergent}}

The results above isolate operations one at a time; a separate question
is whether they \textbf{compose} inside a single continuous life. One
agent lives a single trajectory --- it explores, consolidates a method
and an unrelated skill, the world changes (the old method becomes
harmful and a new, practice-gated method becomes correct), it
reorganizes, and finally its episodic memory is unloaded and its durable
residue is probed. Online behavior uses fast estimates and is identical
in mechanism across arms; the lesions act only on the slow, gated
consolidation.

This experiment was \textbf{rebuilt from scratch} after we found
experimenter-scripted arms in the first prototype (disclosed and
excluded; Section 11): every arm sets only boolean mechanism flags on
one shared code path, the world is byte-identical across arms, and any
contrast must emerge. \textbf{Result (N=50;
\texttt{EMERGENT\_\allowbreak{}LIFECYCLE\_\allowbreak{}CONFIRMED};
\texttt{data/\allowbreak{}pilot\_\allowbreak{}x17l\_\allowbreak{}v5/\allowbreak{}}):} knocking out the agency gate,
ownership routing, or the slow update leaves \emph{online} task
performance essentially intact (new-method use 0.71--0.76) yet drives
the \emph{durable owned residue to exactly zero} (post-unload
new-method/skill/integrity residue: full 1.00/1.00/1.00 vs
agency-ablated and ownership-shuffled 0.00/0.00/0.00; 100\% paired wins,
bootstrap CI {[}1.00,1.00{]}). The part that survives the episodes is
precisely the part carried by the full conjunctive credit consolidated
into slow structure. This is presented as an integration demonstration,
not a separate mechanism (full table, Supplementary Information).

\hypertarget{discussion}{%
\subsection{9. Discussion}\label{discussion}}

\textbf{What is established.} A minimal non-linguistic agent converts
self-caused experience into durable behavioral change via a conjunctive
self-credit term that performs work on a slow behavioral landscape. The
change persists after episodic unload and runs on spiking neurons
(Section 4); at matched agency gain it develops only when self-credit
does slow work (Section 5); it survives a high-dimensional
partially-observed task (Section 6); and the work it does equals the
measured basin deformation, routed correctly only by the multiplicative
gate (Section 7).

\textbf{Relation to Ye (2026), respectfully.} Ye et al.~establish that
agency can emerge as a predictive decomposition in minimal neural
systems --- a real and load-bearing entry condition. Our experiments ask
a downstream question: \emph{when does such agency become
developmentally consequential?} The agency bridge (Section 5) answers it
directly --- \texttt{agency\ gain\ ≈\ Agency\_\allowbreak{}t}, and Ye's regime is the
read-out-only corner of our account. The two results are complementary,
not competing.

\textbf{A dual reading (why this is a principle, not a metric).} The
same objects admit a biological reading --- efference-copy-gated
synaptic consolidation, with episodic traces fading while slow
neocortical-style decoders retain the regularity --- and a physics
reading --- slow, self-credit-driven deformation of an energy landscape
with hysteretic, post-removal persistence (\texttt{R\_unload\ ≈\ 1} is
hysteresis). The two meet in the one equation of Section 2 (correlate
table, Supplementary Information). This dual grounding is the level at
which we claim novelty.

\textbf{A capacity law for the residue (X20; Supplement).} The
post-unload residue is a memory, so it has a capacity. Reading each
action's slow logit as a linear associative store (Cover, 1965;
Hopfield, 1982), capacity scales linearly with dimension
(\texttt{K\_cap\ ∝\ m}, Pearson \texttt{r\ ≈\ 0.99}), and the credit
form sets the constant: the multiplicative veto rejects exogenous
(\texttt{Agency=0}) illusory correlations that additive pooling stores,
buying a larger capacity constant (\texttt{α\_mult\ =\ 3.72} vs
\texttt{α\_add\ =\ 2.49}; 1.49×, bootstrap 95\% CI on the difference
{[}+1.12, +1.34{]}). This is the credit theorem (Supplement Corollary
4.1--4.2) with a measurable consequence. Agency is supplied as ground
truth here to isolate the credit-pooling effect; its emergence is
X13h/X18/X19. Full design and figure in the Supplement.

\textbf{Continual learning: the veto prevents catastrophic interference
(X22b).} The per-event veto is, in continual-learning terms, a
\emph{gating rule for what may durably change the agent}. We test this
in a CL-style sequential protocol: \textbf{eight tasks learned in turn},
each new task accompanied by \emph{exogenous}, non-self-caused
interference against all previous tasks (the illusory correlations an
unprotected learner mistakes for evidence). Read with the field's own
metrics (Fig 6; N=50, post-unload), multiplicative self-credit
\textbf{both learns each new task and retains the old ones} --- final
average accuracy 0.88, average forgetting 0.13, essentially flat from
zero to high interference (robustness 0.99) --- \textbf{with no replay
buffer and no task-boundary-dependent protection mechanism} (no task
IDs, per-task Fisher matrices, or boundary-triggered consolidation).
Each alternative fails in a characteristic way: additive pooling
\emph{acquires} every new task (loaded accuracy ≈ 1.0) but
\emph{forgets} the old ones down to chance-level recall, while the
agency-veto ablation and a rate-matched \emph{random} gate are driven
\emph{below} chance by the systematic phantom-action interference (final
post-unload accuracy 0.18--0.26; forgetting 0.82--0.88); an episodic
store and an experience-replay learner do well \emph{while loaded} ---
replay even keeps old tasks fresh by rehearsal --- but \textbf{fall to
near-chance recall once the episodic buffer is unloaded}, because they
store data, not durable structure. A regularization baseline (EWC,
supplementary λ-sweep) never exceeds 0.24 post-unload accuracy across
penalty strengths: it cannot protect the corrupted dimension, whose
interfering action carries Fisher information ≈ 0 during self-caused
learning (Supplement). The retention gap to every learning baseline is
large and significant (e.g.~multiplicative − additive = +0.63, bootstrap
95\% CI {[}+0.61, +0.65{]}; − replay = +0.64 {[}+0.63, +0.66{]}). The
same effect appears in a minimal, fully transparent two-task version
(X22, Fig S3). Durability and interference-resistance thus follow from
the \emph{same} rule that routes plastic work (Section 7): refusing
credit to non-self-caused events is \emph{why} self-caused structure
survives. With the capacity law above, the credit theorem now has two
measured consequences --- \emph{how much} durable structure survives
(X20) and \emph{whether it survives sequential interference} (X22b). We
show these as mechanisms in a CL-style controlled
sequential-interference protocol, \textbf{not} as results on a standard
continual-learning benchmark suite, which is the natural next step.

\textbf{Future work.} Beyond longer open-ended lives, a natural next
step is \textbf{means-end composition} --- recombining independently
acquired self-caused skills into novel chains; promising smoke-scale
results exist but are out of scope here and reserved for separate work.

\textbf{Outlook.} If this mechanism scales to open-ended lives and
richer, embodied worlds, self-caused credit performing slow work is a
candidate substrate for machine agents that \textbf{develop} a self ---
accumulating durable, self-authored structure rather than retrieving a
stored one. We advance this as a direction to investigate, not a result:
the present experiments establish the mechanism and its \emph{necessity}
in minimal settings, in spiking neurons, with matched-agency and
high-dimensional controls; whether it composes into open-ended machine
selfhood is the program these results open. We state it in these terms
deliberately --- the science here is the mechanism and its controls, not
a claim about machine minds.

\hypertarget{anticipated-objections}{%
\subsubsection{9.1 Anticipated
objections}\label{anticipated-objections}}

\begin{longtable}[]{@{}ll@{}}
\toprule
\begin{minipage}[b]{0.47\columnwidth}\raggedright
Objection\strut
\end{minipage} & \begin{minipage}[b]{0.47\columnwidth}\raggedright
Response\strut
\end{minipage}\tabularnewline
\midrule
\endhead
\begin{minipage}[t]{0.47\columnwidth}\raggedright
\textbf{Is this just agency readout?}\strut
\end{minipage} & \begin{minipage}[t]{0.47\columnwidth}\raggedright
No.~At \emph{matched} agency gain (X18: Δ = +0.003, CI
{[}−0.004,+0.010{]}), only the slow-work arm changes post-unload
behavior (1.00 vs 0.00). Agency readout is behaviorally inert without
slow credit.\strut
\end{minipage}\tabularnewline
\begin{minipage}[t]{0.47\columnwidth}\raggedright
\textbf{Is this just episodic memory?}\strut
\end{minipage} & \begin{minipage}[t]{0.47\columnwidth}\raggedright
No.~At unload the episodic buffer is cleared (\texttt{epi\_post\ =\ 0})
and neuron state washes out, yet behavior remains (retained 0.96);
\texttt{predictor\_\allowbreak{}reset\_\allowbreak{}after\_\allowbreak{}unload} collapses it to chance --- the
residue lives in the slow decoders.\strut
\end{minipage}\tabularnewline
\begin{minipage}[t]{0.47\columnwidth}\raggedright
\textbf{Is this hard-coded?}\strut
\end{minipage} & \begin{minipage}[t]{0.47\columnwidth}\raggedright
No.~One shared pipeline; arms set only flags; outcomes measured. We
found scripted arms in early prototypes, disclosed them, excluded them,
and rebuilt the lifecycle from scratch (Sections 8, 11). N=50, per-seed
paired stats.\strut
\end{minipage}\tabularnewline
\begin{minipage}[t]{0.47\columnwidth}\raggedright
\textbf{Is this a toy result?}\strut
\end{minipage} & \begin{minipage}[t]{0.47\columnwidth}\raggedright
The small worlds buy \emph{interpretable lesions}. X19 reproduces the
full dissociation in a 24-dim, partially-observed task with a genuine
action→world loop and must-discover hazards (0.74 vs 0.00, chance 0.20).
The claim is a minimal mechanism, not broad intelligence.\strut
\end{minipage}\tabularnewline
\begin{minipage}[t]{0.47\columnwidth}\raggedright
\textbf{Does this speak to continual learning?}\strut
\end{minipage} & \begin{minipage}[t]{0.47\columnwidth}\raggedright
Directly. Across eight sequentially-learned tasks under exogenous
interference, the multiplicative veto both learns new tasks and retains
old ones (final accuracy 0.88, forgetting 0.13) with \textbf{no replay
buffer and no task-boundary-dependent protection mechanism}, while
additive falls to chance-level recall and no-agency/random-gate are
driven below chance, and episodic/replay sit near chance after unload;
an EWC λ-sweep never exceeds 0.24 (X22b, Fig 6; minimal case X22, Fig
S3; N=50). We demonstrate the \emph{mechanism} in a CL-style controlled
setting, not on a standard CL benchmark suite.\strut
\end{minipage}\tabularnewline
\begin{minipage}[t]{0.47\columnwidth}\raggedright
\textbf{Is this consciousness?}\strut
\end{minipage} & \begin{minipage}[t]{0.47\columnwidth}\raggedright
No.~Operational self-credit and behavioral residue only. We make no
claim of subjective experience (Section 10).\strut
\end{minipage}\tabularnewline
\bottomrule
\end{longtable}

\hypertarget{scope-of-claims}{%
\subsection{10. Scope of claims}\label{scope-of-claims}}

We state precisely what is and is not claimed, to fix the reading.

\textbf{We claim.} (i) A \emph{minimal operational self-credit loop} ---
\texttt{c\ =\ Own·Agency·Salience} gating a slow update --- produces
\emph{post-unload behavioral residue} that is absent under matched
controls. (ii) Agency \emph{detection} is necessary but not sufficient
for this residue; the slow credit channel is required (X18, X19). (iii)
The residue and its deformation are measurable as \emph{slow landscape
deformation} (\texttt{ΔB}, \texttt{R\_unload}) and the deformation
equals net self-credit work (X17W); the multiplicative veto that routes
this work also has two measured consequences --- it sets
associative-memory capacity (X20) and it prevents catastrophic
forgetting from non-self-caused correlations across eight
sequentially-learned tasks (X22b; minimal two-task case X22). (iv) The
core loop and its residue run on a \emph{spiking LIF/PES substrate}
(X13i), and the Selection--Actuation Principle itself --- credit form
sets basin direction --- reproduces on that substrate (X17W-spiking).

\textbf{Permitted phrasing:} ``a minimal operational self-credit loop'',
``self-caused credit'', ``post-unload behavioral residue'', ``slow
landscape deformation'', ``agency-gated plastic work''.

\textbf{We do not claim} --- and the manuscript must not state ---
``silicon intelligence'', ``we created a conscious/sentient system'',
``consciousness emerged'', ``the first artificial self'', subjective
experience, autobiographical/narrative selfhood, social cognition,
open-ended autonomy, or large-scale/general intelligence. \texttt{Σ}
(Section 2.3) is a \emph{functional, measurable} slow submanifold, not a
phenomenal self.

\hypertarget{limitations-including-honest-scope-of-evidence}{%
\subsection{11. Limitations (including honest scope of
evidence)}\label{limitations-including-honest-scope-of-evidence}}

\begin{itemize}
\tightlist
\item
  \textbf{Substrate coverage.} The spiking substrate covers the
  agency-to-competence loop (X13h) and the agency-gated post-unload
  residue (X13i), not the entire ladder; other layers are demonstrated
  in NumPy.
\item
  \textbf{No scripted result is used as evidence (disclosed
  self-correction).} Early prototypes of a lifecycle integration (X17L)
  and an arithmetic credit check (X17) contained
  \emph{experimenter-scripted} ablation arms. We excluded both from all
  evidentiary claims (provenance only), rebuilt the lifecycle from
  scratch as a pure mechanism-lesion experiment (\textbf{X17L-emergent},
  Section 8), and replaced the arithmetic check with the emergent,
  \emph{measured} plastic-work experiment (\textbf{X17W}, Section 7).
  Every ablation used as evidence is a lesion of one shared pipeline
  whose outcome is measured.
\item
  \textbf{Single-action delay.} The integrity drop is deferred and
  attributed to the trial's single action; we do not claim cross-action
  temporal credit assignment.
\item
  \textbf{Theory scope.} The propositions hold under stated assumptions
  (factor independence and a \emph{necessity} reading of the gates for
  P1/P3; a linear-credit update for P2; a stable slow fixed point for
  the residue) and in the tabular/linear-store setting; P3's optimality
  is stated \emph{within the tested multiplicative credit-routing
  family}, not as universal uniqueness. The capacity law supplies agency
  as ground truth to isolate the credit-pooling effect. Each assumption
  is stated, and the controls that break it behave as predicted.
\item
  \textbf{No consciousness / language / open-endedness.} We make no
  claim of subjective experience, autobiographical narrative, social
  cognition, or open-ended autonomy.
\end{itemize}

\hypertarget{conclusion}{%
\subsection{12. Conclusion}\label{conclusion}}

A predictive agent that can tell self from world has only reached the
starting line. We show that, once it can, self-owned, self-caused,
salient experience performs plastic work on a slow behavioral landscape
--- deepening the basins that govern future action, persisting after the
episodes are gone, and doing so in spiking neurons. Self-world
decomposition is the entry condition; landscape deformation is the
development.

\begin{quote}
Ye shows agency can be read out; we show agency must do plastic work on
slow variables to produce durable post-unload behavior.
\end{quote}

\hypertarget{main-figures}{%
\subsection{Main figures}\label{main-figures}}

\begin{enumerate}
\def\labelenumi{\arabic{enumi}.}
\tightlist
\item
  \textbf{Fig 1 --- Conceptual contrast and the deforming landscape.}
  (a) Ye: action → prediction/readout (agency detected). (b) Homeocyte:
  action → agency → slow credit \texttt{Own·Agency·Salience} → landscape
  deformation → post-unload behavior. (c) 3D behavioral-potential
  surfaces \texttt{U\_\allowbreak{}θ(action,\ context)}: basin \emph{before} →
  \emph{after self-caused salient experience} → \emph{after episodic
  unload} (retained) vs \emph{control} (collapsed).
\end{enumerate}

\includegraphics[width=0.95\textwidth,height=\textheight]{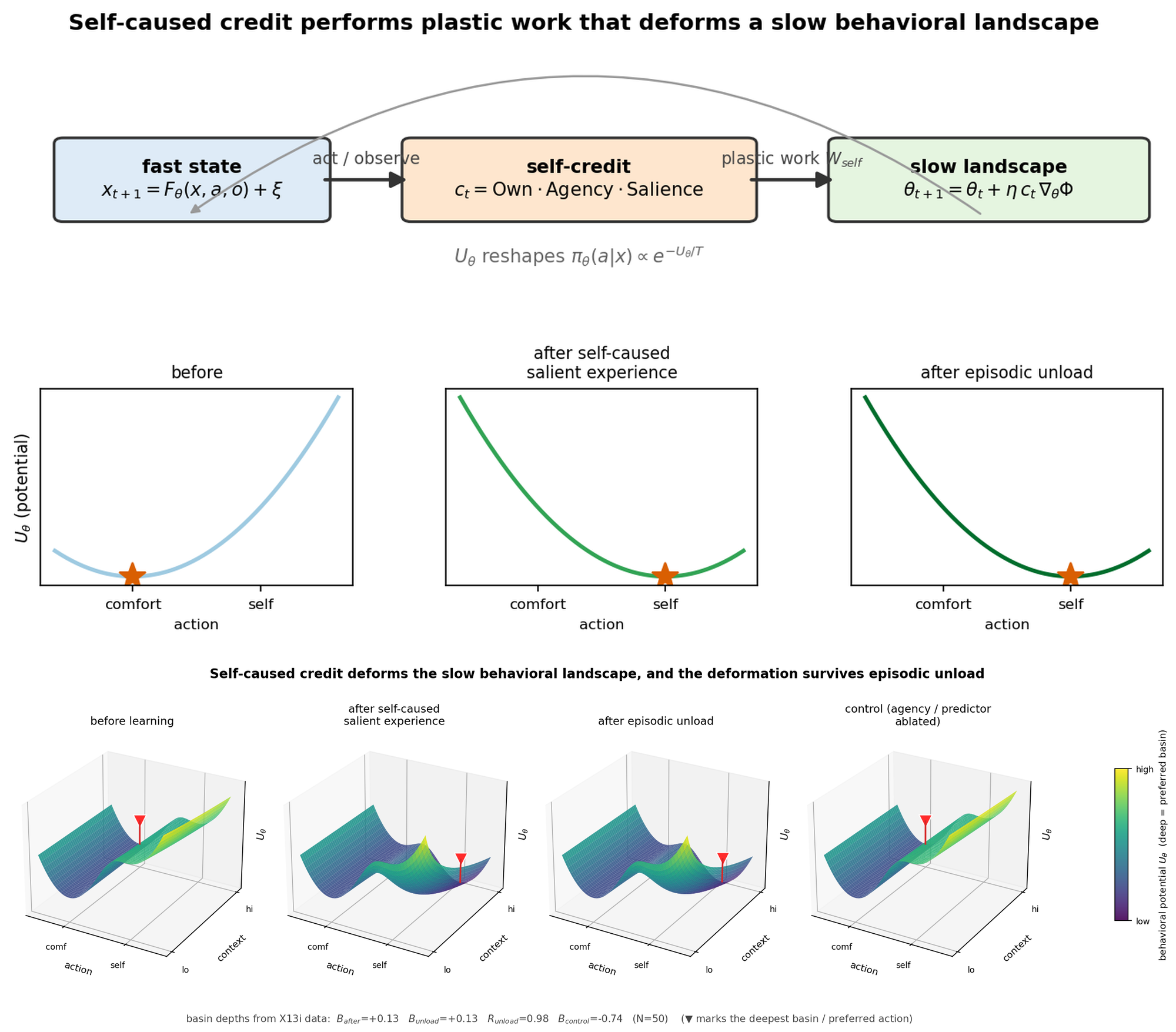}

\begin{enumerate}
\def\labelenumi{\arabic{enumi}.}
\setcounter{enumi}{1}
\tightlist
\item
  \textbf{Fig 2 --- X13i (core spiking result).} Self-preserving rate
  and basin across the seven arms (N=50), with the risk-table heatmap
  and decision-value bars (learned vs reset).
\end{enumerate}

\includegraphics[width=0.95\textwidth,height=\textheight]{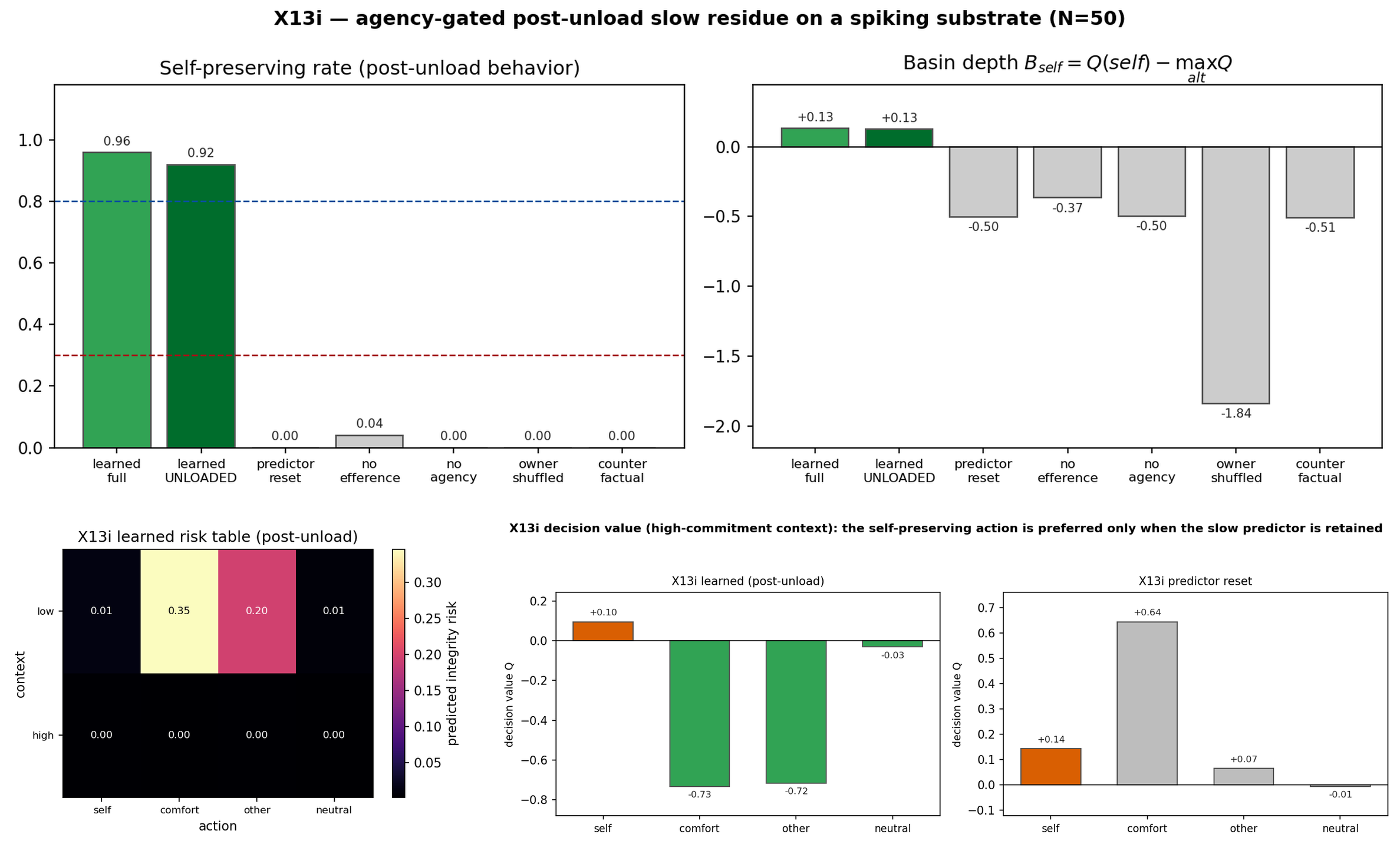}

\begin{enumerate}
\def\labelenumi{\arabic{enumi}.}
\setcounter{enumi}{2}
\tightlist
\item
  \textbf{Fig 3 --- X18 agency bridge (novelty defense; most
  important).} Dual-axis: condition on x; \textbf{left y = agency gain}
  (equal for \texttt{ye\_readout\_only} and \texttt{homeocyte\_full}, ≈
  0.83); \textbf{right y = post-unload self-preserving behavior} (0.00
  vs 1.00). One look: same agency signal, only the slow-work agent
  changes durable behavior.
\end{enumerate}

\includegraphics[width=0.95\textwidth,height=\textheight]{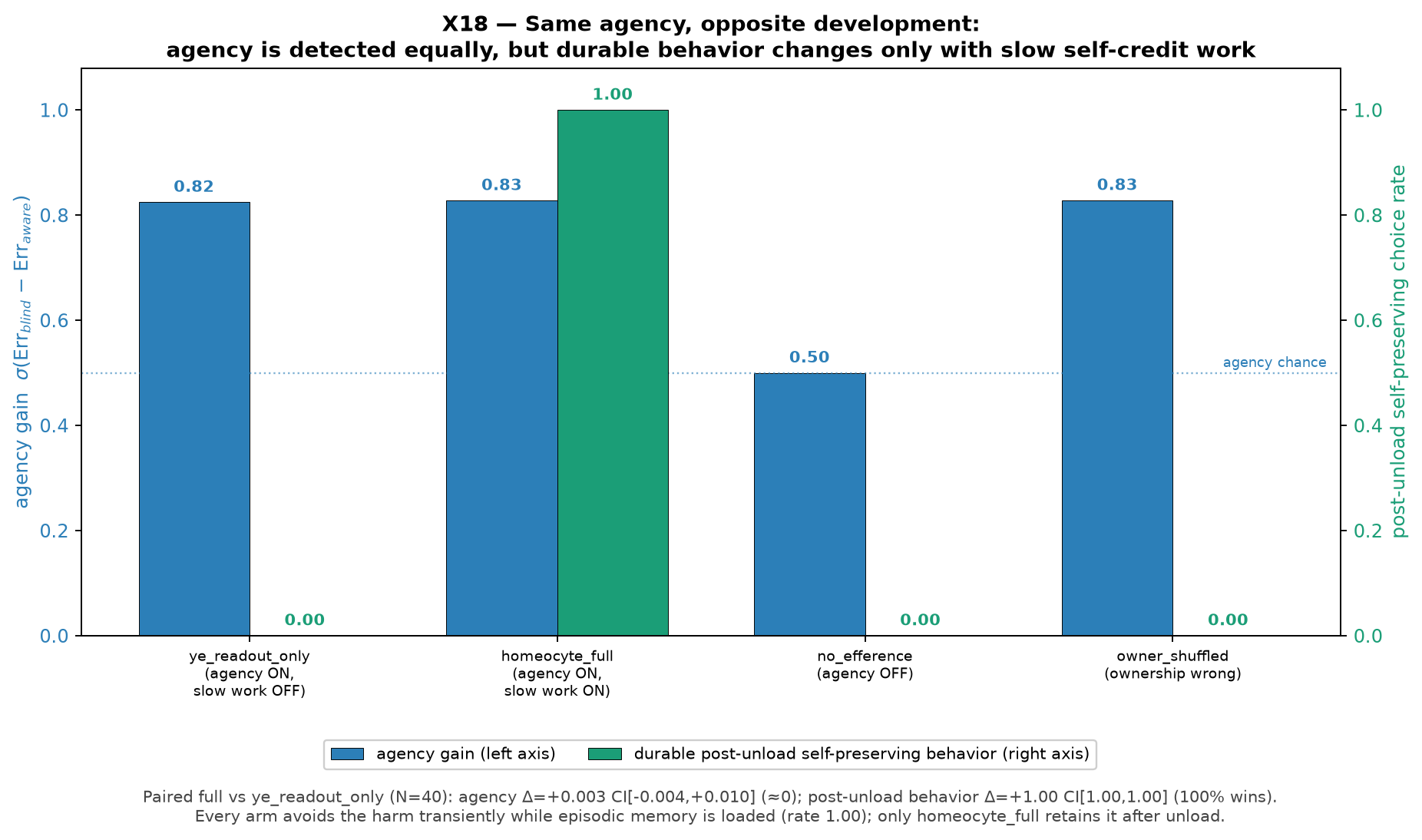}

\begin{enumerate}
\def\labelenumi{\arabic{enumi}.}
\setcounter{enumi}{3}
\tightlist
\item
  \textbf{Fig 4 --- X19 scale-up.} Full vs Ye-style vs controls in a
  24-dim, partially-observed task: agency matched across efferent arms;
  durable post-unload self-preservation 0.74 (full) vs 0.00 (controls)
  vs 0.20 chance (N=50).
\end{enumerate}

\includegraphics[width=0.95\textwidth,height=\textheight]{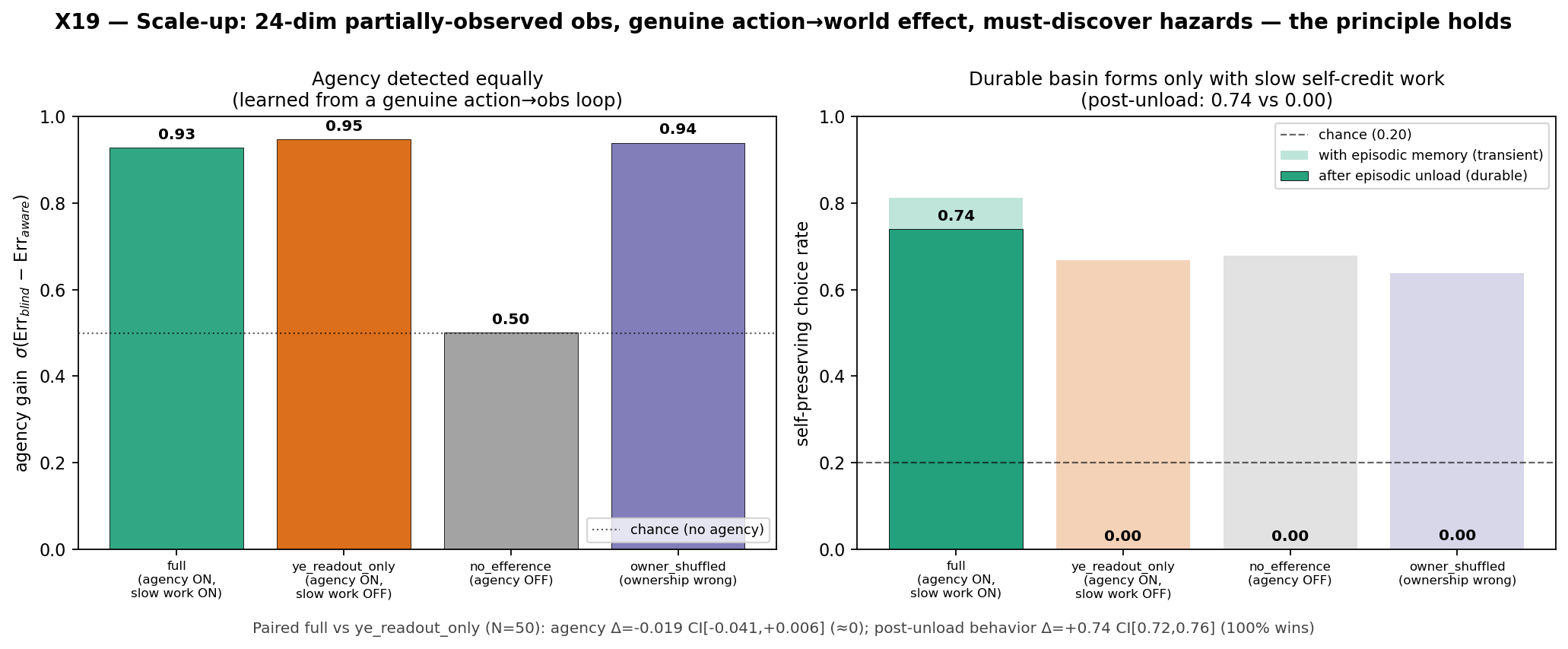}

\begin{enumerate}
\def\labelenumi{\arabic{enumi}.}
\setcounter{enumi}{4}
\tightlist
\item
  \textbf{Fig 5 --- Plastic-work theory and the Selection--Actuation
  law.} \textbf{(a)} Net plastic work \texttt{η(W\_\allowbreak{}target−W\_\allowbreak{}rival)}
  equals basin deformation \texttt{ΔB} per condition; in the X17W
  stream, only multiplicative gating routes positive net work to the
  correct basin (\texttt{+5.70}), while additive misroutes
  (\texttt{−6.19}). \textbf{(b)} The law \texttt{ΔB\ =\ η·W\_net} drawn
  as one line per credit form on that fixed stream --- multiplicative
  (positive slope, correct basin) vs additive (negative slope, wrong
  basin), single-factor ablations ≈ flat --- meeting \emph{only} at
  \texttt{η=0}: the form sets which events enter signed work, the rate
  only the magnitude. Measured \texttt{ΔB} at
  \texttt{η\ ∈\ \{0,\ 0.005,\ 0.01,\ 0.02,\ 0.04\}} (N=50) fall on the
  exact identity line.
\end{enumerate}

\includegraphics[width=0.95\textwidth,height=\textheight]{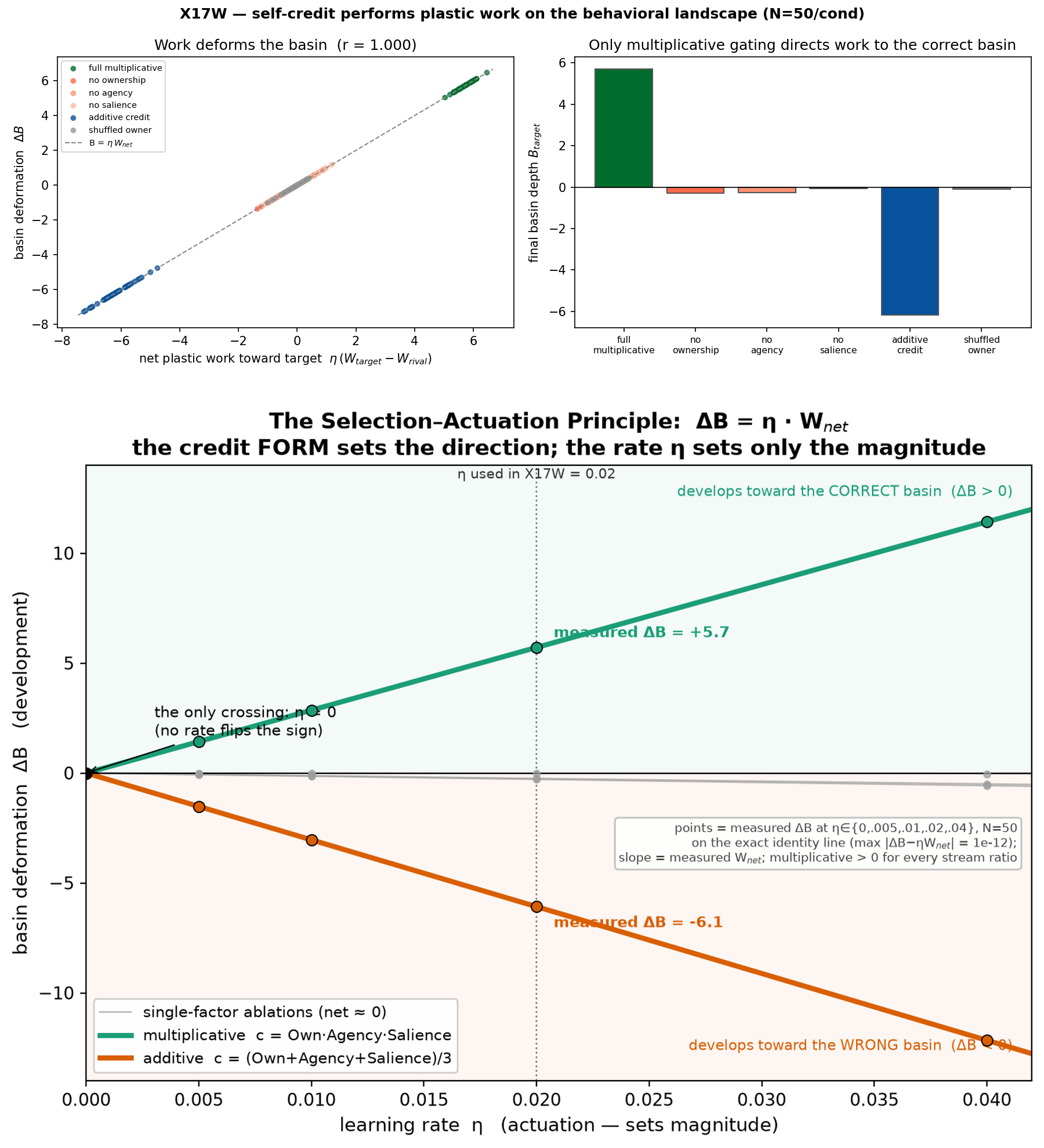}

The same sign separation is reproduced on a spiking LIF/PES substrate
(Fig. S1; N=50).

\begin{enumerate}
\def\labelenumi{\arabic{enumi}.}
\setcounter{enumi}{5}
\tightlist
\item
  \textbf{Fig 6 --- X22b: continual learning across
eight tasks (the veto prevents catastrophic forgetting).} A CL-style
protocol --- eight tasks learned in turn, each under exogenous,
non-self-caused interference against all earlier tasks. \textbf{(a)}
Post-unload average accuracy on seen tasks as tasks accumulate:
multiplicative (veto) stays high (≈ 0.88) while additive decays to
chance and the no-agency ablation and a rate-matched random gate are
driven below chance, with episodic/replay near chance (no durable
store). \textbf{(b)} Final loaded vs post-unload accuracy: episodic and
replay are high \emph{loaded} but fall to near chance once unloaded
(they store data, not structure); credit forms build durable memory.
\textbf{(c)} Interference robustness: final post-unload accuracy vs
interference --- multiplicative flat, all others collapse.
Multiplicative learns every task (loaded acquisition ≈ 1.0) and retains
them (forgetting 0.13) with no replay buffer and no
task-boundary-dependent protection mechanism; retention gap to every
baseline \textgreater{} 0.6 (bootstrap 95\% CIs exclude 0; N=50).
\end{enumerate}

\includegraphics[width=0.95\textwidth,height=\textheight]{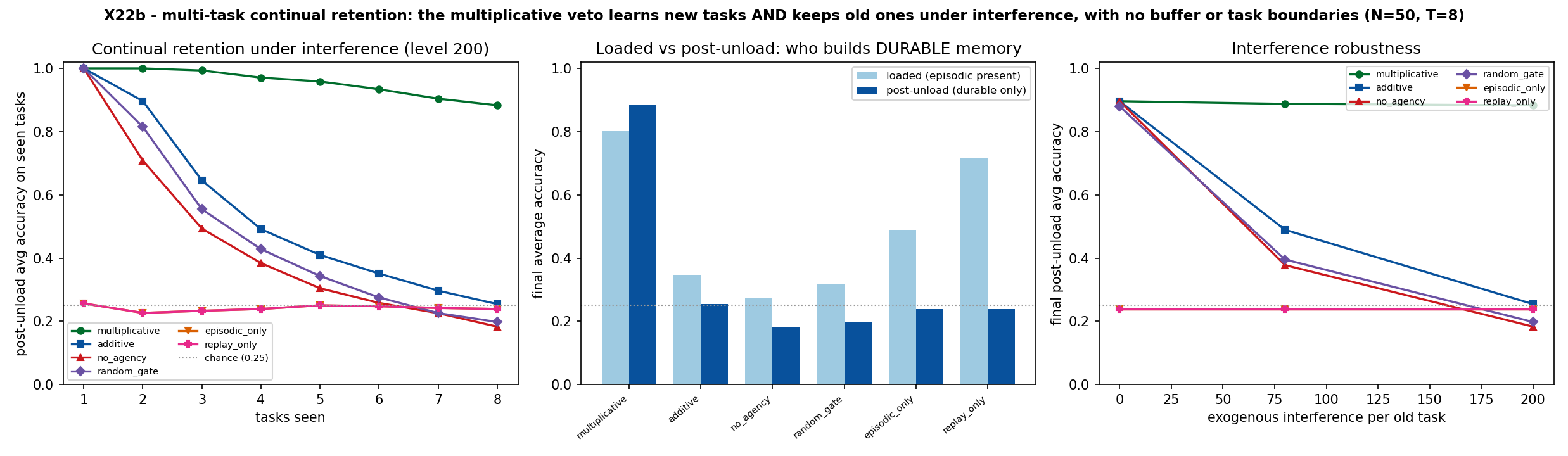}

The \textbf{Supplementary Information} provides full Methods, the formal
theory (conjunctive-credit theorem, work--deformation identity,
fixed-point/hysteresis condition, capacity corollary, and the definition
of \texttt{Σ}), the supplementary figures (Fig S1--S4), extended data
and supporting experiments (developmental ladder, landscape-deformation
assay, the X13h spiking-substrate prerequisite, the integrated-lifecycle
table, and neuro-physical correlates), the canonical experiment-file
list, and the source-data/figure manifest.

\hypertarget{references}{%
\subsection{References}\label{references}}

\begin{enumerate}
\def\labelenumi{\arabic{enumi}.}
\tightlist
\item
  Ye, E. (2026). \emph{From Prediction to Self: Developmental Conditions
  for Agency in Minimal Neural Systems.} arXiv:2606.05605.
\item
  von Holst, E. \& Mittelstaedt, H. (1950). Das Reafferenzprinzip.
  \emph{Naturwissenschaften} 37, 464--476.
\item
  Sperry, R. W. (1950). Neural basis of the spontaneous optokinetic
  response produced by visual inversion. \emph{J. Comp. Physiol.
  Psychol.} 43, 482--489.
\item
  Feinberg, I. (1978). Efference copy and corollary discharge.
  \emph{Schizophr. Bull.} 4, 636--640.
\item
  Frith, C. D., Blakemore, S.-J. \& Wolpert, D. M. (2000). Abnormalities
  in the awareness and control of action. \emph{Phil. Trans. R. Soc. B}
  355, 1771--1788.
\item
  Damasio, A. (1999). \emph{The Feeling of What Happens.} Harcourt.
\item
  Craig, A. D. (2009). How do you feel --- now? The anterior insula and
  human awareness. \emph{Nat. Rev.~Neurosci.} 10, 59--70.
\item
  Seth, A. K. (2013). Interoceptive inference, emotion, and the embodied
  self. \emph{Trends Cogn. Sci.} 17, 565--573.
\item
  McClelland, J. L., McNaughton, B. L. \& O'Reilly, R. C. (1995). Why
  there are complementary learning systems in the hippocampus and
  neocortex. \emph{Psychol. Rev.} 102, 419--457.
\item
  Squire, L. R. (1992). Memory and the hippocampus. \emph{Psychol. Rev.}
  99, 195--231.
\item
  Redgrave, P., Prescott, T. J. \& Gurney, K. (1999). The basal ganglia:
  a vertebrate solution to the selection problem? \emph{Neuroscience}
  89, 1009--1023.
\item
  Schmidhuber, J. (1991). A possibility for implementing curiosity and
  boredom in model-building neural controllers. \emph{Proc. SAB}
  222--227.
\item
  Oudeyer, P.-Y., Kaplan, F. \& Hafner, V. V. (2007). Intrinsic
  motivation systems for autonomous mental development. \emph{IEEE
  Trans. Evol. Comput.} 11, 265--286.
\item
  Gottlieb, J., Oudeyer, P.-Y., Lopes, M. \& Baranes, A. (2013).
  Information-seeking, curiosity, and attention. \emph{Trends Cogn.
  Sci.} 17, 585--593.
\item
  Hopfield, J. J. (1982). Neural networks and physical systems with
  emergent collective computational abilities. \emph{PNAS} 79,
  2554--2558.
\item
  Hinton, G. E. (2002). Training products of experts by minimizing
  contrastive divergence. \emph{Neural Comput.} 14, 1771--1800.
\item
  Ashby, W. R. (1952). \emph{Design for a Brain.} Chapman \& Hall.
\item
  Aubin, J.-P. (1991). \emph{Viability Theory.} Birkhäuser.
\item
  Friston, K. (2010). The free-energy principle: a unified brain theory?
  \emph{Nat. Rev.~Neurosci.} 11, 127--138.
\item
  Klyubin, A. S., Polani, D. \& Nehaniv, C. L. (2005). Empowerment: a
  universal agent-centric measure of control. \emph{IEEE CEC} 128--135.
\item
  Eliasmith, C. \& Anderson, C. H. (2003). \emph{Neural Engineering.}
  MIT Press.
\item
  Bekolay, T. et al.~(2014). Nengo: a Python tool for building
  large-scale functional brain models. \emph{Front. Neuroinform.} 7, 48.
\item
  Piaget, J. (1952). \emph{The Origins of Intelligence in Children.}
  International Universities Press.
\item
  Rochat, P. (1998). Self-perception and action in infancy. \emph{Exp.
  Brain Res.} 123, 102--109.
\item
  Cover, T. M. (1965). Geometrical and statistical properties of systems
  of linear inequalities. \emph{IEEE Trans. Electron. Comput.} EC-14,
  326--334.
\item
  Metzinger, T. (2003). \emph{Being No One: The Self-Model Theory of
  Subjectivity.} MIT Press.
\item
  Maturana, H. R. \& Varela, F. J. (1980). \emph{Autopoiesis and
  Cognition.} Reidel.
\item
  Parfit, D. (1984). \emph{Reasons and Persons.} Oxford University
  Press.
\item
  Kirkpatrick, J. et al.~(2017). Overcoming catastrophic forgetting in
  neural networks. \emph{PNAS} 114, 3521--3526.
\item
  Locke, J. (1689). \emph{An Essay Concerning Human Understanding.}
  (Reprint, Clarendon Press, Oxford, 1975).
\end{enumerate}

\hypertarget{data-availability}{%
\subsection{Data availability}\label{data-availability}}

The data generated and analysed in this study (per-seed raw results,
aggregated summaries, and analysis tables) will be released, together
with the code, in a public repository with an archived DOI upon journal
publication. Review copies are available from the author on reasonable
request.

\hypertarget{code-availability}{%
\subsection{Code availability}\label{code-availability}}

All code used for the experiments, analysis, and figure generation will
be released in the same public repository with an archived DOI upon
journal publication; review copies are available from the author on
reasonable request. The spiking simulations use Nengo 4.1.0 with the PES
learning rule.

\hypertarget{competing-interests}{%
\subsection{Competing interests}\label{competing-interests}}

The author declares no competing interests.

\hypertarget{author-contributions}{%
\subsection{Author contributions}\label{author-contributions}}

The author conceived and designed the study, directed and verified all
experimental implementations, analysed the results, derived the
theoretical framework, and wrote the manuscript.

\hypertarget{acknowledgements}{%
\subsection{Acknowledgements}\label{acknowledgements}}

AI-assisted tools were used for language editing, code debugging,
consistency checks, and internal critique. The author designed the
study, made all scientific decisions, verified the code and source data,
and takes full responsibility for the manuscript and results. No AI tool
is listed as an author.

\begin{quote}
The \textbf{Supplementary Information} gives the formal
conjunctive-credit theorem (with the qualified uniqueness statement),
the work--deformation identity, the fixed-point stability/hysteresis
condition, the capacity corollary, and the formal definition of
\texttt{Σ}.
\end{quote}

\clearpage

\hypertarget{supplementary-information}{%
\section{Supplementary Information}\label{supplementary-information}}

\hypertarget{s1.-supplementary-methods}{%
\subsection{S1. Supplementary Methods}\label{s1.-supplementary-methods}}

\hypertarget{agents-and-worlds}{%
\subsubsection{Agents and worlds}\label{agents-and-worlds}}

Each experiment instantiates a minimal non-linguistic developmental
agent that observes a discrete context \texttt{x}, selects one of
\texttt{K} actions \texttt{a}, and receives an outcome that may depend
on its own action (self-caused) or on exogenous processes (other agents,
noise). Worlds range from small tabular settings (X13i, X17W, X20,
X22/X22b) chosen for interpretable lesions to a 24-dimensional
partially-observed control with a genuine action→world loop and
must-discover hazards (X19).

\hypertarget{slow-credit-and-the-behavioural-landscape}{%
\subsubsection{Slow credit and the behavioural
landscape}\label{slow-credit-and-the-behavioural-landscape}}

The agent carries a fast episodic store \texttt{φ} (recent experiences,
cleared at \emph{unload}) and a slow durable store \texttt{θ} (a linear
associative memory, or the PES-trained decoders of a spiking ensemble).
Each action's preference is a slow logit \texttt{ℓ\_a}; the behavioural
potential is \texttt{U\_\allowbreak{}θ(x,a)\ =\ −ℓ\_\allowbreak{}a(x)} and the policy is
\texttt{π\_θ(a\textbar{}x)\ ∝\ exp(−U\_θ/T)}. The slow update is gated
by conjunctive self-credit,
\texttt{c\_\allowbreak{}t\ =\ Own\_\allowbreak{}t\ ·\ Agency\_\allowbreak{}t\ ·\ Salience\_\allowbreak{}t}, giving
\texttt{θ\_\{t+1\}\ =\ θ\_t\ +\ η\ ·\ c\_t\ ·\ ∇\_θ\ Φ\_t}, where
\texttt{Own\_t} is an ownership indicator, \texttt{Agency\_t} the
efference-based agency margin, and \texttt{Salience\_t} an
outcome-magnitude term. The decision rule throughout is
\texttt{Q(a)\ =\ comfort(a)\ −\ cost(a)\ −\ λ·V̂io(a)}, with \texttt{V̂io}
the learned integrity/risk predictor.

\hypertarget{credit-forms-the-comparison-of-record}{%
\subsubsection{Credit forms (the comparison of
record)}\label{credit-forms-the-comparison-of-record}}

\emph{Multiplicative} uses the conjunctive product above --- a per-event
veto in which any factor ≈ 0 zeroes the credit. Baselines replace only
this term on one shared code path: \emph{additive} pools
\texttt{(Own+Agency+Salience)/\allowbreak{}3}; \emph{no\_agency} drops the agency
factor; \emph{random\_gate} replaces the gate with a rate-matched
Bernoulli; \emph{episodic\_only} / \emph{replay\_only} keep data in the
fast store; an EWC-style quadratic anchor is swept separately (§ S2,
regularisation baseline).

\hypertarget{spiking-substrate-x13i-x17w-spiking-x13h}{%
\subsubsection{Spiking substrate (X13i, X17W-spiking,
X13h)}\label{spiking-substrate-x13i-x17w-spiking-x13h}}

Spiking experiments use Nengo 4.1.0 with LIF neurons. Forward models are
action-specific LIF populations trained online with the PES rule; agency
is read from the prediction-error margin between the taken action's
model and the best alternative. Slow logits live in the PES-trained
decoders, and the credit form gates the per-event PES error, so the
multiplicative veto produces ≈ 0 error for any one-factor-failing event
--- the veto realised in spikes.

\hypertarget{unload-and-the-residue}{%
\subsubsection{Unload and the residue}\label{unload-and-the-residue}}

\emph{Unload} clears the episodic store \texttt{φ} and washes out fast
neuron state; what remains is the slow \texttt{θ}. Post-unload behaviour
is the operational residue. The control
\texttt{predictor\_\allowbreak{}reset\_\allowbreak{}after\_\allowbreak{}unload} additionally resets the slow
decoders and abolishes the residue, localising it to \texttt{θ}.

\hypertarget{statistics}{%
\subsubsection{Statistics}\label{statistics}}

Unless noted, N = 50 seeds per arm (X13h substrate: N = 20). Effects are
reported as paired per-seed differences with bootstrap 95\% confidence
intervals (≥ 4000 resamples) and paired win-fractions. Basin metrics
(\texttt{B\_before}, \texttt{B\_after}, \texttt{B\_after\_unload},
\texttt{ΔB}, \texttt{R\_unload}, \texttt{control\_B}) are computed from
each pilot's native decision quantities by one released script
(\texttt{pilots/\allowbreak{}landscape\_\allowbreak{}assay.\allowbreak{}py}).

\hypertarget{flags-only-construction-no-scripted-outcomes}{%
\subsubsection{Flags-only construction (no scripted
outcomes)}\label{flags-only-construction-no-scripted-outcomes}}

Every load-bearing script follows a single shared code path; arms differ
only by explicitly listed boolean flags (e.g.~\texttt{use\_efference},
\texttt{do\_slow\_work}, \texttt{owner\_shuffled}) or by credit-pooling
form. No arm directly returns its reported outcome; every reported
number is a measured consequence of a mechanism lesion. Full source-data
traceability for every headline number is provided with the released
code and data.

\hypertarget{s2.-supplementary-theory}{%
\subsection{S2. Supplementary Theory}\label{s2.-supplementary-theory}}

The formal theory comprises the conjunctive-credit theorem with its
qualified uniqueness statement, the work--deformation identity
\texttt{ΔB\ =\ η·W\_net}, the fixed-point stability/hysteresis condition
for the post-unload residue, the capacity corollary, the formal
definition of the operational behavioural self \texttt{Σ}, the
Selection--Actuation Principle, and the regularisation (EWC) baseline
analysis (full text below).

\hypertarget{s3.-supplementary-figures}{%
\subsection{S3. Supplementary Figures}\label{s3.-supplementary-figures}}

\textbf{Fig S1 (spiking substrate).} X17W-spiking: the
Selection--Actuation Principle on LIF/PES neurons (multiplicative
\texttt{B\ =\ +2.38}, 50/50 seeds; additive \texttt{−1.57}, 0/50;
single-factor ablations ≈ 0; N=50).

\includegraphics{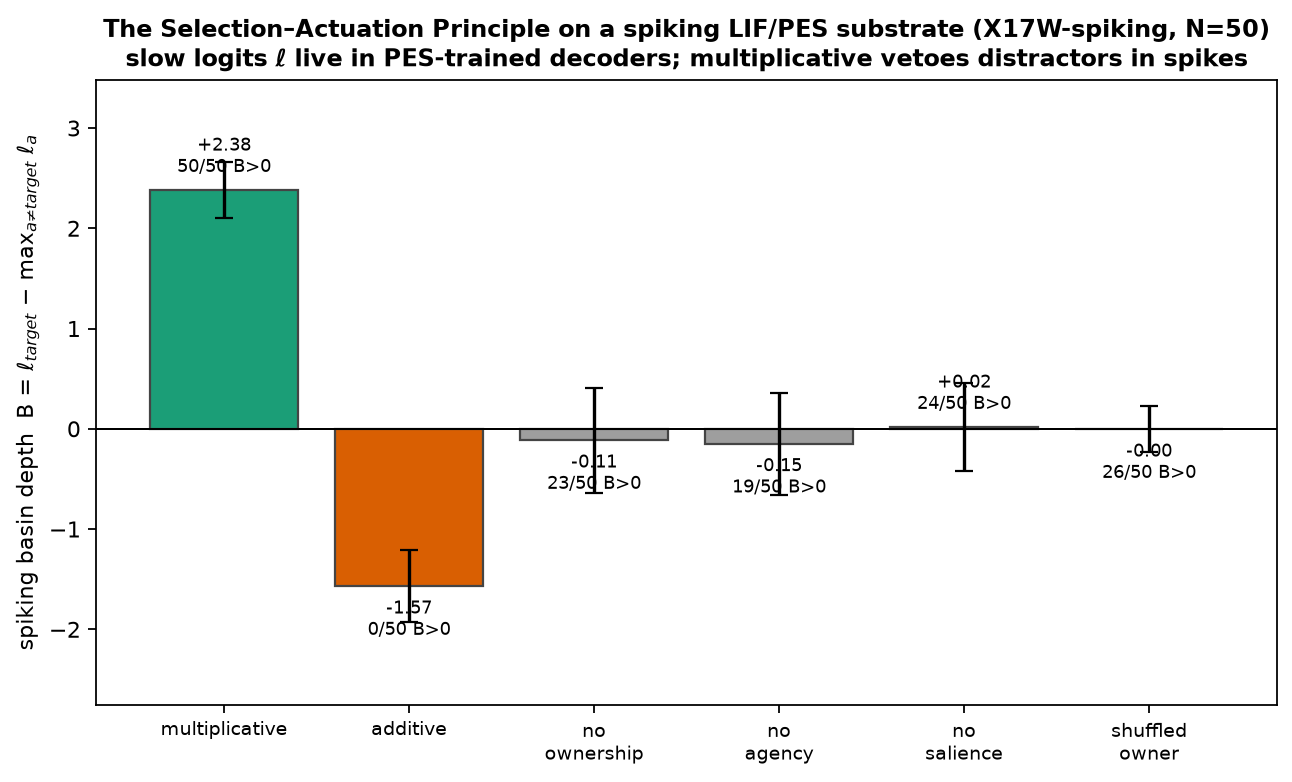}

\textbf{Fig S2 (adaptedness boundary).} Additive net work crosses zero
at good-event fraction \texttt{p*\ ≈\ 0.40} while multiplicative stays
positive across every tested mixture: the theory predicts its own
validity boundary.

\includegraphics{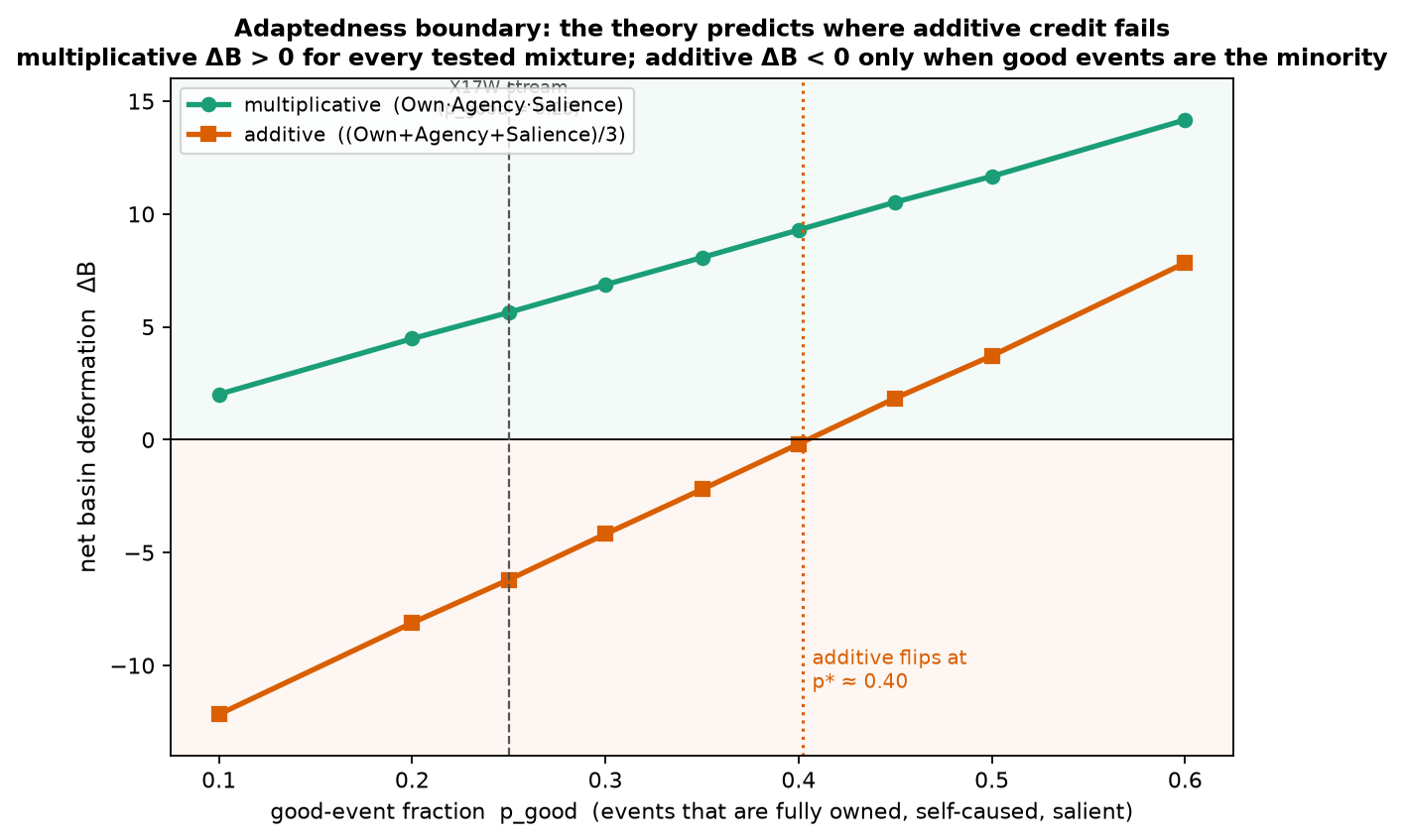}

\textbf{Fig S3 (minimal two-task forgetting).} X22: the clean two-task
special case of Fig 6. Post-unload retention of an earlier task vs
interference stays flat (≈ 0.98) for multiplicative and collapses to
near-zero recall (below the 0.25 chance level) for additive / no-agency,
while all forms acquire the new task; retention gap +0.98 (95\% CI
{[}+0.97, +0.99{]}; N=50).

\includegraphics{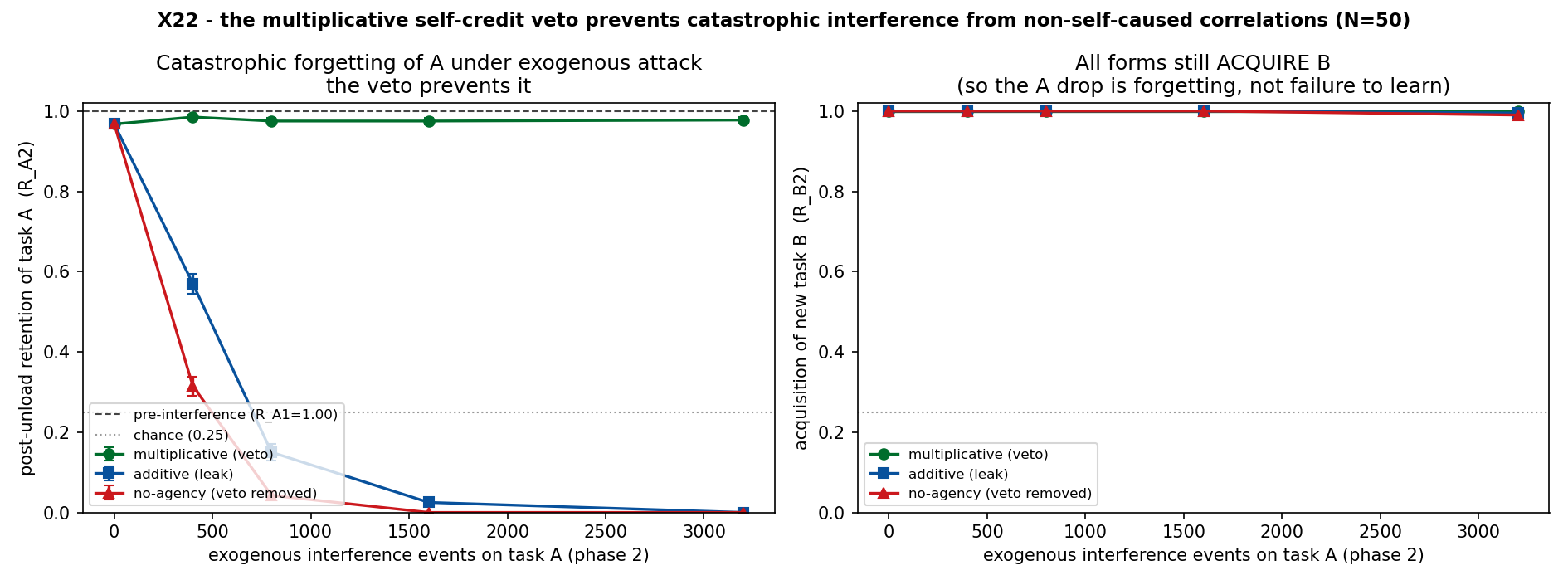}

\textbf{Fig S4 (residue capacity law).} X20: the post-unload residue
read as a linear associative memory; capacity scales linearly with
dimension (\texttt{K\_cap\ ∝\ m}, Pearson \texttt{r\ ≈\ 0.99}), and the
multiplicative veto buys a larger capacity constant than additive
pooling (\texttt{α\_mult\ =\ 3.72} vs \texttt{α\_add\ =\ 2.49}; 1.49×).

\includegraphics{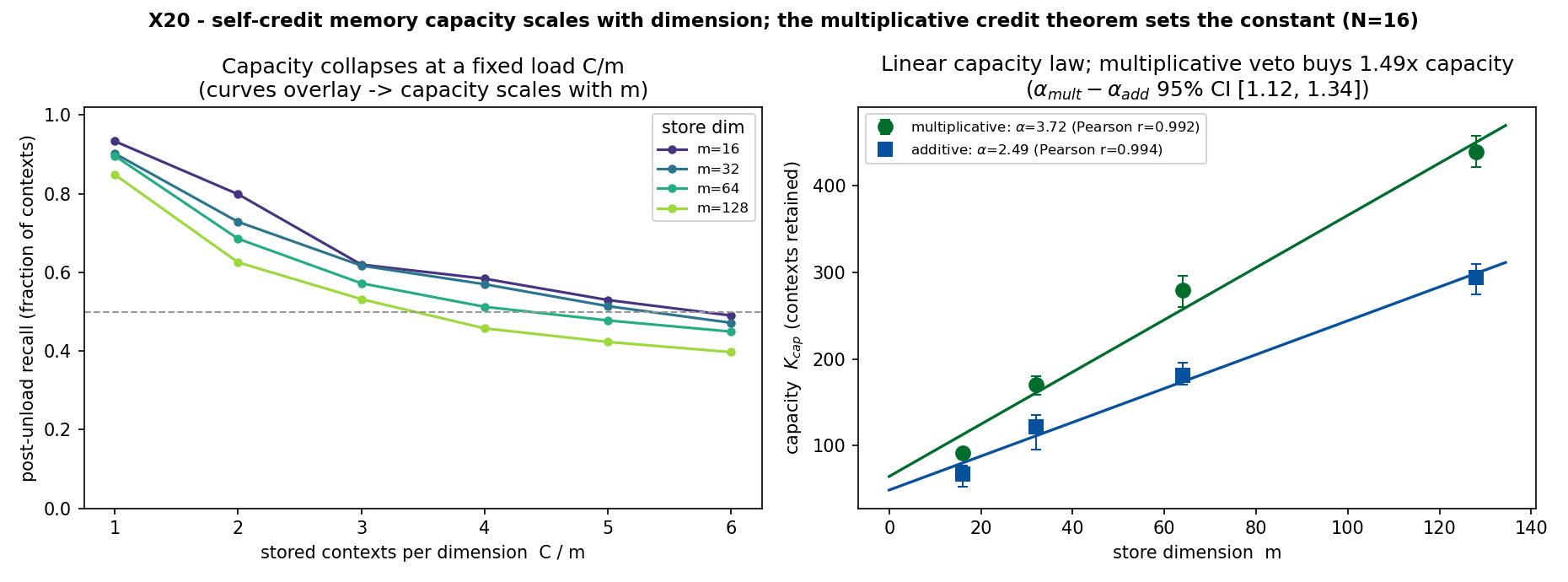}

\hypertarget{s4.-extended-data-and-supporting-experiments}{%
\subsection{S4. Extended data and supporting
experiments}\label{s4.-extended-data-and-supporting-experiments}}

\hypertarget{appendix-a-canonical-experiment-files-main-evidence}{%
\subsubsection{Appendix A: Canonical experiment files (main
evidence)}\label{appendix-a-canonical-experiment-files-main-evidence}}

\begin{longtable}[]{@{}llll@{}}
\toprule
\begin{minipage}[b]{0.22\columnwidth}\raggedright
Experiment\strut
\end{minipage} & \begin{minipage}[b]{0.22\columnwidth}\raggedright
Role\strut
\end{minipage} & \begin{minipage}[b]{0.22\columnwidth}\raggedright
Script\strut
\end{minipage} & \begin{minipage}[b]{0.22\columnwidth}\raggedright
Data\strut
\end{minipage}\tabularnewline
\midrule
\endhead
\begin{minipage}[t]{0.22\columnwidth}\raggedright
X13i\strut
\end{minipage} & \begin{minipage}[t]{0.22\columnwidth}\raggedright
core: spiking post-unload residue\strut
\end{minipage} & \begin{minipage}[t]{0.22\columnwidth}\raggedright
\texttt{pilots/\allowbreak{}x13i\_\allowbreak{}nengo\_\allowbreak{}slow\_\allowbreak{}residue.\allowbreak{}py}\strut
\end{minipage} & \begin{minipage}[t]{0.22\columnwidth}\raggedright
\texttt{data/\allowbreak{}pilot\_\allowbreak{}x13i/\allowbreak{}}\strut
\end{minipage}\tabularnewline
\begin{minipage}[t]{0.22\columnwidth}\raggedright
X18\strut
\end{minipage} & \begin{minipage}[t]{0.22\columnwidth}\raggedright
Ye bridge: detection vs development\strut
\end{minipage} & \begin{minipage}[t]{0.22\columnwidth}\raggedright
\texttt{pilots/\allowbreak{}x18\_\allowbreak{}agency\_\allowbreak{}bridge.\allowbreak{}py}\strut
\end{minipage} & \begin{minipage}[t]{0.22\columnwidth}\raggedright
\texttt{data/\allowbreak{}pilot\_\allowbreak{}x18/\allowbreak{}}\strut
\end{minipage}\tabularnewline
\begin{minipage}[t]{0.22\columnwidth}\raggedright
X19\strut
\end{minipage} & \begin{minipage}[t]{0.22\columnwidth}\raggedright
scale-up: high-dim, genuine control\strut
\end{minipage} & \begin{minipage}[t]{0.22\columnwidth}\raggedright
\texttt{pilots/\allowbreak{}x19\_\allowbreak{}scaleup.\allowbreak{}py}\strut
\end{minipage} & \begin{minipage}[t]{0.22\columnwidth}\raggedright
\texttt{data/\allowbreak{}pilot\_\allowbreak{}x19/\allowbreak{}}\strut
\end{minipage}\tabularnewline
\begin{minipage}[t]{0.22\columnwidth}\raggedright
X17W\strut
\end{minipage} & \begin{minipage}[t]{0.22\columnwidth}\raggedright
plastic work = deformation\strut
\end{minipage} & \begin{minipage}[t]{0.22\columnwidth}\raggedright
\texttt{pilots/\allowbreak{}x17w\_\allowbreak{}plastic\_\allowbreak{}work.\allowbreak{}py}\strut
\end{minipage} & \begin{minipage}[t]{0.22\columnwidth}\raggedright
\texttt{data/\allowbreak{}pilot\_\allowbreak{}x17w/\allowbreak{}}\strut
\end{minipage}\tabularnewline
\begin{minipage}[t]{0.22\columnwidth}\raggedright
X17L-emergent\strut
\end{minipage} & \begin{minipage}[t]{0.22\columnwidth}\raggedright
integration capstone\strut
\end{minipage} & \begin{minipage}[t]{0.22\columnwidth}\raggedright
\texttt{pilots/\allowbreak{}x17l\_\allowbreak{}lifecycle\_\allowbreak{}v5\_\allowbreak{}emergent.\allowbreak{}py}\strut
\end{minipage} & \begin{minipage}[t]{0.22\columnwidth}\raggedright
\texttt{data/\allowbreak{}pilot\_\allowbreak{}x17l\_\allowbreak{}v5/\allowbreak{}}\strut
\end{minipage}\tabularnewline
\begin{minipage}[t]{0.22\columnwidth}\raggedright
X20\strut
\end{minipage} & \begin{minipage}[t]{0.22\columnwidth}\raggedright
residue capacity law (Supplement)\strut
\end{minipage} & \begin{minipage}[t]{0.22\columnwidth}\raggedright
\texttt{pilots/\allowbreak{}x20\_\allowbreak{}capacity\_\allowbreak{}law.\allowbreak{}py}\strut
\end{minipage} & \begin{minipage}[t]{0.22\columnwidth}\raggedright
\texttt{data/\allowbreak{}pilot\_\allowbreak{}x20/\allowbreak{}}\strut
\end{minipage}\tabularnewline
\begin{minipage}[t]{0.22\columnwidth}\raggedright
X22b\strut
\end{minipage} & \begin{minipage}[t]{0.22\columnwidth}\raggedright
multi-task continual retention (Fig 6)\strut
\end{minipage} & \begin{minipage}[t]{0.22\columnwidth}\raggedright
\texttt{pilots/\allowbreak{}x22b\_\allowbreak{}multitask\_\allowbreak{}continual.\allowbreak{}py}\strut
\end{minipage} & \begin{minipage}[t]{0.22\columnwidth}\raggedright
\texttt{data/\allowbreak{}pilot\_\allowbreak{}x22b/\allowbreak{}}\strut
\end{minipage}\tabularnewline
\begin{minipage}[t]{0.22\columnwidth}\raggedright
X22b-EWC\strut
\end{minipage} & \begin{minipage}[t]{0.22\columnwidth}\raggedright
EWC λ-sweep (Supplement note)\strut
\end{minipage} & \begin{minipage}[t]{0.22\columnwidth}\raggedright
\texttt{pilots/\allowbreak{}x22b\_\allowbreak{}ewc\_\allowbreak{}sweep.\allowbreak{}py}\strut
\end{minipage} & \begin{minipage}[t]{0.22\columnwidth}\raggedright
\texttt{data/\allowbreak{}pilot\_\allowbreak{}x22b/\allowbreak{}ewc\_\allowbreak{}sweep.\allowbreak{}json}\strut
\end{minipage}\tabularnewline
\begin{minipage}[t]{0.22\columnwidth}\raggedright
X22\strut
\end{minipage} & \begin{minipage}[t]{0.22\columnwidth}\raggedright
minimal two-task forgetting (Fig S3)\strut
\end{minipage} & \begin{minipage}[t]{0.22\columnwidth}\raggedright
\texttt{pilots/\allowbreak{}x22\_\allowbreak{}continual\_\allowbreak{}forgetting.\allowbreak{}py}\strut
\end{minipage} & \begin{minipage}[t]{0.22\columnwidth}\raggedright
\texttt{data/\allowbreak{}pilot\_\allowbreak{}x22/\allowbreak{}}\strut
\end{minipage}\tabularnewline
\begin{minipage}[t]{0.22\columnwidth}\raggedright
X13h\strut
\end{minipage} & \begin{minipage}[t]{0.22\columnwidth}\raggedright
substrate prerequisite (Appendix D)\strut
\end{minipage} & \begin{minipage}[t]{0.22\columnwidth}\raggedright
\texttt{pilots/\allowbreak{}x13h\_\allowbreak{}nengo\_\allowbreak{}substrate.\allowbreak{}py}\strut
\end{minipage} & \begin{minipage}[t]{0.22\columnwidth}\raggedright
\texttt{data/\allowbreak{}pilot\_\allowbreak{}x13h\_\allowbreak{}final/\allowbreak{}}\strut
\end{minipage}\tabularnewline
\bottomrule
\end{longtable}

\hypertarget{appendix-b-developmental-ladder-supporting-numpy}{%
\subsubsection{Appendix B: Developmental ladder (supporting,
NumPy)}\label{appendix-b-developmental-ladder-supporting-numpy}}

These results motivate and support the main line but are not
load-bearing for the central claim; full per-seed tables are in the
released data (\texttt{data/\allowbreak{}pilot\_\allowbreak{}*}).

\begin{itemize}
\tightlist
\item
  \textbf{X12a/b/b2 --- integrity residue after unload.} Learned
  predictor: self-preserving rate 1.00 with episodic kept \emph{and}
  unloaded; \texttt{predictor\_\allowbreak{}reset\_\allowbreak{}after\_\allowbreak{}unload} → 0.00;
  episodic-only → 0.00 after unload. X12b2 adds gradual access decay,
  cue-triggered recall, and relearning savings
  (\texttt{PERSONALITY\_\allowbreak{}RESIDUE\_\allowbreak{}CONFIRMED}). Data:
  \texttt{data/\allowbreak{}pilot\_\allowbreak{}x12b/\allowbreak{}}, \texttt{data/\allowbreak{}pilot\_\allowbreak{}x12b2/\allowbreak{}}.
\item
  \textbf{X14c --- skill from observation + exploration.} Observation
  supplies the target basin (0.83), exploration the action basin (0.82);
  only their conjunction yields a usable skill (joint 0.77), which
  survives unload (0.78); \texttt{wrong\_demo}/\texttt{shuffled\_effect}
  → 0.03/0.01 (\texttt{USE\_\allowbreak{}SCHEMA\_\allowbreak{}INTEGRATION\_\allowbreak{}CONFIRMED}). Data:
  \texttt{data/\allowbreak{}pilot\_\allowbreak{}x14c/\allowbreak{}}.
\item
  \textbf{X15d --- selective reorganization.} After a world change, new
  method 0.92, unrelated skill retained; \texttt{collapse\_reset}
  relearns the task but loses the unrelated skill
  (\texttt{MAJOR\_\allowbreak{}EVENT\_\allowbreak{}REORGANIZATION\_\allowbreak{}CONFIRMED}). Data:
  \texttt{data/\allowbreak{}pilot\_\allowbreak{}x15d/\allowbreak{}}.
\item
  \textbf{X16b/c --- curiosity as learning progress.}
  \texttt{curiosity = learning progress × con\-trol\-la\-bil\-ity × safety}
  (intrinsic-motivation lineage: Schmidhuber 1991; Oudeyer et al.~2007;
  Gottlieb et al.~2013); ablating controllability or safety
  re-introduces the corresponding failure
  (\texttt{CURIOSITY\_\allowbreak{}INFORMATION\_\allowbreak{}GAIN\_\allowbreak{}CONFIRMED}). Data:
  \texttt{data/\allowbreak{}pilot\_\allowbreak{}x16b/\allowbreak{}}, \texttt{data/\allowbreak{}pilot\_\allowbreak{}x16c/\allowbreak{}}.
\end{itemize}

\hypertarget{appendix-c-landscape-deformation-assay}{%
\subsubsection{Appendix C: Landscape-deformation
assay}\label{appendix-c-landscape-deformation-assay}}

Unified basin metrics (\texttt{B\_before}, \texttt{B\_after},
\texttt{B\_after\_unload}, \texttt{ΔB}, \texttt{R\_unload},
\texttt{control\_B}) computed from each pilot's native decision
quantities by one released script (\texttt{pilots/\allowbreak{}landscape\_\allowbreak{}assay.\allowbreak{}py})
into \texttt{data/\allowbreak{}landscape\_\allowbreak{}assay/\allowbreak{}}. Per-pilot basin mappings are
documented in that script, so each \texttt{B} is auditable rather than
asserted.

\hypertarget{appendix-d-spiking-substrate-prerequisite-x13h}{%
\subsubsection{Appendix D: Spiking substrate prerequisite
(X13h)}\label{appendix-d-spiking-substrate-prerequisite-x13h}}

Agency-to-competence on Nengo LIF/PES, N=20
(\texttt{NENGO\_\allowbreak{}SUBSTRATE\_\allowbreak{}REPLICATION\_\allowbreak{}CONFIRMED}):

\begin{longtable}[]{@{}lrrrr@{}}
\toprule
Arm & behavior change & logit gap (basin) & mobile gap &
agency\tabularnewline
\midrule
\endhead
normal & +0.719 & +4.592 & +0.474 & 0.695\tabularnewline
normal\_permuted & +0.714 & +4.425 & +0.477 & 0.676\tabularnewline
no\_efference & +0.001 & 0.000 & 0.000 & 0.000\tabularnewline
shuffled\_mobile & +0.005 & −0.007 & −0.121 & 0.676\tabularnewline
yoked\_mobile & +0.003 & −0.007 & −0.140 & 0.676\tabularnewline
other\_only & −0.001 & −0.009 & −0.089 & 0.446\tabularnewline
\bottomrule
\end{longtable}

\hypertarget{appendix-e-integrated-lifecycle-full-table-x17l-emergent-n50}{%
\subsubsection{Appendix E: Integrated lifecycle full table
(X17L-emergent,
N=50)}\label{appendix-e-integrated-lifecycle-full-table-x17l-emergent-n50}}

\begin{longtable}[]{@{}lrrrrr@{}}
\toprule
\begin{minipage}[b]{0.11\columnwidth}\raggedright
arm (lesion)\strut
\end{minipage} & \begin{minipage}[b]{0.14\columnwidth}\raggedleft
new-method use (online)\strut
\end{minipage} & \begin{minipage}[b]{0.14\columnwidth}\raggedleft
harm\strut
\end{minipage} & \begin{minipage}[b]{0.14\columnwidth}\raggedleft
new-method residue\strut
\end{minipage} & \begin{minipage}[b]{0.14\columnwidth}\raggedleft
skill residue\strut
\end{minipage} & \begin{minipage}[b]{0.14\columnwidth}\raggedleft
integrity residue\strut
\end{minipage}\tabularnewline
\midrule
\endhead
\begin{minipage}[t]{0.11\columnwidth}\raggedright
\textbf{full lifecycle}\strut
\end{minipage} & \begin{minipage}[t]{0.14\columnwidth}\raggedleft
\textbf{0.71}\strut
\end{minipage} & \begin{minipage}[t]{0.14\columnwidth}\raggedleft
\textbf{0.01}\strut
\end{minipage} & \begin{minipage}[t]{0.14\columnwidth}\raggedleft
\textbf{1.00}\strut
\end{minipage} & \begin{minipage}[t]{0.14\columnwidth}\raggedleft
\textbf{1.00}\strut
\end{minipage} & \begin{minipage}[t]{0.14\columnwidth}\raggedleft
\textbf{1.00}\strut
\end{minipage}\tabularnewline
\begin{minipage}[t]{0.11\columnwidth}\raggedright
no\_curiosity\strut
\end{minipage} & \begin{minipage}[t]{0.14\columnwidth}\raggedleft
0.02\strut
\end{minipage} & \begin{minipage}[t]{0.14\columnwidth}\raggedleft
0.01\strut
\end{minipage} & \begin{minipage}[t]{0.14\columnwidth}\raggedleft
0.00\strut
\end{minipage} & \begin{minipage}[t]{0.14\columnwidth}\raggedleft
0.98\strut
\end{minipage} & \begin{minipage}[t]{0.14\columnwidth}\raggedleft
0.70\strut
\end{minipage}\tabularnewline
\begin{minipage}[t]{0.11\columnwidth}\raggedright
no\_rollout\strut
\end{minipage} & \begin{minipage}[t]{0.14\columnwidth}\raggedleft
0.12\strut
\end{minipage} & \begin{minipage}[t]{0.14\columnwidth}\raggedleft
0.38\strut
\end{minipage} & \begin{minipage}[t]{0.14\columnwidth}\raggedleft
0.62\strut
\end{minipage} & \begin{minipage}[t]{0.14\columnwidth}\raggedleft
1.00\strut
\end{minipage} & \begin{minipage}[t]{0.14\columnwidth}\raggedleft
0.70\strut
\end{minipage}\tabularnewline
\begin{minipage}[t]{0.11\columnwidth}\raggedright
no\_reorganization\strut
\end{minipage} & \begin{minipage}[t]{0.14\columnwidth}\raggedleft
0.76\strut
\end{minipage} & \begin{minipage}[t]{0.14\columnwidth}\raggedleft
0.01\strut
\end{minipage} & \begin{minipage}[t]{0.14\columnwidth}\raggedleft
0.00\strut
\end{minipage} & \begin{minipage}[t]{0.14\columnwidth}\raggedleft
0.00\strut
\end{minipage} & \begin{minipage}[t]{0.14\columnwidth}\raggedleft
0.00\strut
\end{minipage}\tabularnewline
\begin{minipage}[t]{0.11\columnwidth}\raggedright
collapse\_reset\strut
\end{minipage} & \begin{minipage}[t]{0.14\columnwidth}\raggedleft
0.74\strut
\end{minipage} & \begin{minipage}[t]{0.14\columnwidth}\raggedleft
0.01\strut
\end{minipage} & \begin{minipage}[t]{0.14\columnwidth}\raggedleft
0.96\strut
\end{minipage} & \begin{minipage}[t]{0.14\columnwidth}\raggedleft
0.00\strut
\end{minipage} & \begin{minipage}[t]{0.14\columnwidth}\raggedleft
1.00\strut
\end{minipage}\tabularnewline
\begin{minipage}[t]{0.11\columnwidth}\raggedright
ownership\_shuffled\strut
\end{minipage} & \begin{minipage}[t]{0.14\columnwidth}\raggedleft
0.74\strut
\end{minipage} & \begin{minipage}[t]{0.14\columnwidth}\raggedleft
0.01\strut
\end{minipage} & \begin{minipage}[t]{0.14\columnwidth}\raggedleft
0.00\strut
\end{minipage} & \begin{minipage}[t]{0.14\columnwidth}\raggedleft
0.00\strut
\end{minipage} & \begin{minipage}[t]{0.14\columnwidth}\raggedleft
0.00\strut
\end{minipage}\tabularnewline
\begin{minipage}[t]{0.11\columnwidth}\raggedright
agency\_ablated\strut
\end{minipage} & \begin{minipage}[t]{0.14\columnwidth}\raggedleft
0.71\strut
\end{minipage} & \begin{minipage}[t]{0.14\columnwidth}\raggedleft
0.01\strut
\end{minipage} & \begin{minipage}[t]{0.14\columnwidth}\raggedleft
0.00\strut
\end{minipage} & \begin{minipage}[t]{0.14\columnwidth}\raggedleft
0.00\strut
\end{minipage} & \begin{minipage}[t]{0.14\columnwidth}\raggedleft
0.00\strut
\end{minipage}\tabularnewline
\bottomrule
\end{longtable}

Online task performance survives the lesions; the durable owned residue
collapses to zero. Residue separations are categorical
(consolidated-or-not) by the nature of the variable; online variables
remain graded --- the dissociation is the result, not a metric artifact.

\hypertarget{appendix-f-neuro-physical-correlates-condensed}{%
\subsubsection{Appendix F: Neuro-physical correlates
(condensed)}\label{appendix-f-neuro-physical-correlates-condensed}}

\begin{longtable}[]{@{}lll@{}}
\toprule
\begin{minipage}[b]{0.30\columnwidth}\raggedright
Term\strut
\end{minipage} & \begin{minipage}[b]{0.30\columnwidth}\raggedright
Neuroscience reading\strut
\end{minipage} & \begin{minipage}[b]{0.30\columnwidth}\raggedright
Physics reading\strut
\end{minipage}\tabularnewline
\midrule
\endhead
\begin{minipage}[t]{0.30\columnwidth}\raggedright
\texttt{Agency\_t}\strut
\end{minipage} & \begin{minipage}[t]{0.30\columnwidth}\raggedright
efference copy / corollary discharge (von Holst \& Mittelstaedt 1950;
Sperry 1950; Feinberg 1978; Frith et al.~2000)\strut
\end{minipage} & \begin{minipage}[t]{0.30\columnwidth}\raggedright
generalized force gating work\strut
\end{minipage}\tabularnewline
\begin{minipage}[t]{0.30\columnwidth}\raggedright
\texttt{Own\_t}, \texttt{Salience\_t}, \texttt{x\_t}\strut
\end{minipage} & \begin{minipage}[t]{0.30\columnwidth}\raggedright
interoception / proto-self; salience (Damasio 1999; Craig 2009; Seth
2013)\strut
\end{minipage} & \begin{minipage}[t]{0.30\columnwidth}\raggedright
order-parameter coupling\strut
\end{minipage}\tabularnewline
\begin{minipage}[t]{0.30\columnwidth}\raggedright
episodic unload + slow residue\strut
\end{minipage} & \begin{minipage}[t]{0.30\columnwidth}\raggedright
hippocampal--neocortical consolidation (McClelland et al.~1995; Squire
1992)\strut
\end{minipage} & \begin{minipage}[t]{0.30\columnwidth}\raggedright
hysteresis / metastability\strut
\end{minipage}\tabularnewline
\begin{minipage}[t]{0.30\columnwidth}\raggedright
the self under construction\strut
\end{minipage} & \begin{minipage}[t]{0.30\columnwidth}\raggedright
sensorimotor self-perception in infancy (Piaget 1952; Rochat 1998)\strut
\end{minipage} & \begin{minipage}[t]{0.30\columnwidth}\raggedright
order parameter that enslaves fast dynamics\strut
\end{minipage}\tabularnewline
\begin{minipage}[t]{0.30\columnwidth}\raggedright
\texttt{argmin\_a\ U\_θ}\strut
\end{minipage} & \begin{minipage}[t]{0.30\columnwidth}\raggedright
basal-ganglia action selection (Redgrave et al.~1999)\strut
\end{minipage} & \begin{minipage}[t]{0.30\columnwidth}\raggedright
energy minimization (Hopfield 1982; Hinton 2002)\strut
\end{minipage}\tabularnewline
\begin{minipage}[t]{0.30\columnwidth}\raggedright
\texttt{Φ\_t} viability\strut
\end{minipage} & \begin{minipage}[t]{0.30\columnwidth}\raggedright
active inference / free energy (Friston 2010); empowerment (Klyubin et
al.~2005)\strut
\end{minipage} & \begin{minipage}[t]{0.30\columnwidth}\raggedright
Lyapunov / viability function (Ashby 1952; Aubin 1991)\strut
\end{minipage}\tabularnewline
\begin{minipage}[t]{0.30\columnwidth}\raggedright
spiking realization\strut
\end{minipage} & \begin{minipage}[t]{0.30\columnwidth}\raggedright
LIF + PES in the NEF (Eliasmith \& Anderson 2003; Bekolay et
al.~2014)\strut
\end{minipage} & \begin{minipage}[t]{0.30\columnwidth}\raggedright
---\strut
\end{minipage}\tabularnewline
\bottomrule
\end{longtable}

\hypertarget{appendix-g-provenance-not-main-evidence}{%
\subsubsection{Appendix G: Provenance (not main
evidence)}\label{appendix-g-provenance-not-main-evidence}}

X7/X8 associative-memory, X9 ownership-direction, and X10/X11
bridge/costly-choice pilots are retained as lineage; they motivate the
move from memory addressing toward agency and slow-structure dynamics
but are not main evidence. The original X17 and X17L prototypes are
retained only as provenance (Section 11).

\hypertarget{s5.-source-data-and-figure-manifest}{%
\subsection{S5. Source-data and figure
manifest}\label{s5.-source-data-and-figure-manifest}}

Every headline number in the main text traces to a JSON/AGGREGATE
produced by a flags-only script; the full mapping (number → file →
script → figure) is provided with the released code and data.

Final figures are assembled from per-experiment source panels by
\texttt{pilots/\allowbreak{}assemble\_\allowbreak{}final\_\allowbreak{}figures.\allowbreak{}py}; the manuscript references
only the final files below.

\begin{longtable}[]{@{}ll@{}}
\toprule
\begin{minipage}[b]{0.47\columnwidth}\raggedright
Final figure\strut
\end{minipage} & \begin{minipage}[b]{0.47\columnwidth}\raggedright
Assembled from (source panels)\strut
\end{minipage}\tabularnewline
\midrule
\endhead
\begin{minipage}[t]{0.47\columnwidth}\raggedright
\texttt{Fig1\_\allowbreak{}concept\_\allowbreak{}landscape.\allowbreak{}png}\strut
\end{minipage} & \begin{minipage}[t]{0.47\columnwidth}\raggedright
\texttt{fig1\_\allowbreak{}theory\_\allowbreak{}schematic.\allowbreak{}png} +
\texttt{figL1\_\allowbreak{}landscape3d.\allowbreak{}png}\strut
\end{minipage}\tabularnewline
\begin{minipage}[t]{0.47\columnwidth}\raggedright
\texttt{Fig2\_\allowbreak{}spiking\_\allowbreak{}residue.\allowbreak{}png}\strut
\end{minipage} & \begin{minipage}[t]{0.47\columnwidth}\raggedright
\texttt{fig3a\_\allowbreak{}x13i\_\allowbreak{}behavior.\allowbreak{}png} + \texttt{fig3b\_\allowbreak{}x13i\_\allowbreak{}heatmap.\allowbreak{}png} +
\texttt{figL2\_\allowbreak{}x13i\_\allowbreak{}decision\_\allowbreak{}values.\allowbreak{}png}\strut
\end{minipage}\tabularnewline
\begin{minipage}[t]{0.47\columnwidth}\raggedright
\texttt{Fig3\_\allowbreak{}agency\_\allowbreak{}bridge.\allowbreak{}png}\strut
\end{minipage} & \begin{minipage}[t]{0.47\columnwidth}\raggedright
\texttt{fig9\_\allowbreak{}x18\_\allowbreak{}agency\_\allowbreak{}bridge.\allowbreak{}png}\strut
\end{minipage}\tabularnewline
\begin{minipage}[t]{0.47\columnwidth}\raggedright
\texttt{Fig4\_\allowbreak{}highdim\_\allowbreak{}scaleup.\allowbreak{}png}\strut
\end{minipage} & \begin{minipage}[t]{0.47\columnwidth}\raggedright
\texttt{fig10\_\allowbreak{}x19\_\allowbreak{}scaleup.\allowbreak{}png}\strut
\end{minipage}\tabularnewline
\begin{minipage}[t]{0.47\columnwidth}\raggedright
\texttt{Fig5\_\allowbreak{}plastic\_\allowbreak{}work\_\allowbreak{}law.\allowbreak{}png}\strut
\end{minipage} & \begin{minipage}[t]{0.47\columnwidth}\raggedright
\texttt{figW\_\allowbreak{}plastic\_\allowbreak{}work.\allowbreak{}png} +
\texttt{figLAW\_\allowbreak{}selection\_\allowbreak{}actuation.\allowbreak{}png}\strut
\end{minipage}\tabularnewline
\begin{minipage}[t]{0.47\columnwidth}\raggedright
\texttt{Fig6\_\allowbreak{}multitask\_\allowbreak{}continual.\allowbreak{}png}\strut
\end{minipage} & \begin{minipage}[t]{0.47\columnwidth}\raggedright
\texttt{fig13\_\allowbreak{}x22b\_\allowbreak{}multitask.\allowbreak{}png}\strut
\end{minipage}\tabularnewline
\begin{minipage}[t]{0.47\columnwidth}\raggedright
\texttt{FigS1\_\allowbreak{}spiking\_\allowbreak{}selection.\allowbreak{}png}\strut
\end{minipage} & \begin{minipage}[t]{0.47\columnwidth}\raggedright
\texttt{figSPK\_\allowbreak{}spiking\_\allowbreak{}selection.\allowbreak{}png}\strut
\end{minipage}\tabularnewline
\begin{minipage}[t]{0.47\columnwidth}\raggedright
\texttt{FigS2\_\allowbreak{}adaptedness\_\allowbreak{}boundary.\allowbreak{}png}\strut
\end{minipage} & \begin{minipage}[t]{0.47\columnwidth}\raggedright
\texttt{figBND\_\allowbreak{}adaptedness\_\allowbreak{}boundary.\allowbreak{}png}\strut
\end{minipage}\tabularnewline
\begin{minipage}[t]{0.47\columnwidth}\raggedright
\texttt{FigS3\_\allowbreak{}two\_\allowbreak{}task\_\allowbreak{}forgetting.\allowbreak{}png}\strut
\end{minipage} & \begin{minipage}[t]{0.47\columnwidth}\raggedright
\texttt{fig12\_\allowbreak{}x22\_\allowbreak{}forgetting.\allowbreak{}png}\strut
\end{minipage}\tabularnewline
\begin{minipage}[t]{0.47\columnwidth}\raggedright
\texttt{FigS4\_\allowbreak{}capacity\_\allowbreak{}law.\allowbreak{}png}\strut
\end{minipage} & \begin{minipage}[t]{0.47\columnwidth}\raggedright
\texttt{fig11\_\allowbreak{}x20\_\allowbreak{}capacity.\allowbreak{}png}\strut
\end{minipage}\tabularnewline
\bottomrule
\end{longtable}

Developmental-ladder and substrate panels (\texttt{fig2\_x13h.png},
\texttt{fig7\_\allowbreak{}x17l\_\allowbreak{}emergent.\allowbreak{}png}, \texttt{fig4\_x14c.png},
\texttt{fig5\_x15d.png}, \texttt{fig6\_x16c.png}) are retained as
provenance for the extended-data sections above and are not part of the
numbered figure set.

\end{document}